\documentclass{article}

\usepackage{graphicx}
\usepackage{subcaption} %
\usepackage{amsmath}
\usepackage{todonotes}
\usepackage{algorithm}
\usepackage{algpseudocode}

\usepackage[preprint]{neurips_2026}

\usepackage{iftex}

\usepackage{hyperref}       %
\usepackage{url}            %
\usepackage{booktabs}       %
\usepackage{amsfonts}       %
\usepackage{nicefrac}       %
\usepackage{microtype}      %
\usepackage{xcolor}         %

\title{The Ephemeral Cost of Switching Optimizers\\Across LLM Training Stages}

\author{
    Xinyu Zhao \thanks{Work done during an internship at MIT-IBM Watson AI Lab. Correspondence to: \texttt{lxz@mit.edu}}\\
    MIT \\
    \And
    Davis Wertheimer \\
    MIT-IBM Watson AI Lab \\
    \And 
    Ashutosh Sharma \\
    MIT-IBM Watson AI Lab \\
    \AND 
    Rameswar Panda \\
    MIT-IBM Watson AI Lab \\
    \And
    Paul Pu Liang \\
    MIT
}

\begin{document}

\maketitle

\begin{abstract}
  Recent work suggests that switching optimizers between pretraining and subsequent training stages incurs a performance cost, but systematic investigation of this phenomenon is lacking. 
  We study the mismatch between AdamW and Muon optimizers by pretraining 1.2B Llama-3 models from scratch while varying the proportion of updates taken with each optimizer, and similarly interpolating between the two optimizers during continued pretraining and supervised fine-tuning (SFT). Surprisingly, we find little evidence of a meaningful optimizer mismatch penalty: final performance is largely explained by the independent effects of the pretraining and post-training optimizers, with downstream performance generally improving as the proportion of Muon updates during SFT increases.
  Although matching the pretraining optimizer can reduce forgetting during SFT, 
  this does not translate into improved final benchmark performance. We further evaluate AdamW and Muon for task-general SFT of open-weight models, across the Llama and Qwen families up to 14B parameters, where Muon remains competitive with AdamW, despite these models having been pretrained with AdamW. Finally, we investigate whether our findings extend to RL via experiments on our 1.2B models as well as Qwen3-1.7B and 4B, and find that AdamW and Muon achieve comparable final performance in math RLVR. Together, our results suggest that the choice of optimizer for post-training can be made largely independently of the optimizer used during pretraining. Our code is released \href{https://anonymous.4open.science/r/sft-optimizer-BB44}{here}.
\end{abstract}

\section{Introduction}

AdamW \citep{loshchilov2019decoupled} has been the de facto optimizer for training most large language models. However, recent results have shown that Muon \citep{jordan2024muon} can outperform AdamW at scale in LLM pretraining \citep{liu2025muon, khona2026soap}. Yet it remains an open question whether Muon is also a better optimizer for post-training stages, and whether switching optimizers between training stages is detrimental to final performance.
Prior work has suggested such a penalty for using a different optimizer for SFT \citep{qu2026muon}, due to differing regularization effects \citep{liu2026optimizer} or implicit biases \citep{qu2026muon}. It has been unclear whether this effect is limited to greater forgetting during task-specific fine-tuning, or also extends to degraded final benchmark performance (``performance cost") following the larger-scale, multi-task SFT stage typical in foundation model training. In RLVR training stages, where forgetting is largely mitigated during AdamW training due to the nature of on-policy updates \citep{shenfeld2026rls}, prior work finds that Muon may underperform \citep{fan2026rethinkingmuonpretrainingspectral} or be competitive with AdamW \citep{ruan2026doesmuonhelpagentic} in single-seed settings depending on specific implementation and hyperparameter choices.

In this work, we examine whether there is a cost to switching between AdamW and Muon across LLM training stages, when hyperparameters for both optimizers are carefully tuned and across multiple training seeds. We train our own 1.2B Llama-3 models from scratch while varying the proportion of AdamW and Muon updates used during pretraining, continued pretraining, and task-general SFT, allowing us to separate the effect of optimizer mismatch from each optimizer's independent effect, as well as that of the underlying data shift. We additionally evaluate AdamW and Muon for task-general SFT of released open-weight models up to 14B parameters. Finally, we study whether these findings carry over to RLVR by performing GRPO on math with matched and mismatched optimizers on both our own 1.2B pretrained models and Qwen3 models.

Our main findings are as follows. First, after accounting for the marginal effects of the pretraining and post-training optimizers, we find little evidence of an additional performance cost from switching optimizers: interaction effects are small during SFT, and apparent switching costs during continued pretraining largely disappear after tuning the learning rate for each setting. Second, optimizer consistency does reduce forgetting of the pretraining distribution during SFT, but this does not translate into improved downstream benchmark performance. Instead, increasing the proportion of Muon updates during SFT generally improves performance regardless of the pretraining optimizer. Furthermore, across AdamW-pretrained open-weight models from the Llama and Qwen families up to 14B parameters, Muon remains competitive with AdamW for task-general SFT. In math RLVR, we find that Muon and AdamW achieve comparable final downstream performance at their selected learning rates regardless of pretraining optimizer, and forgetting on downstream benchmarks is similarly within noise.

\section{Experimental Setup}

For our 1.2B-parameter experiments, we pre-train Llama-3 transformer models with 32 layers, 24 attention heads with full multi-head-attention, and a hidden dimension of 1536, following the hyperparameter recommendations for this configuration given by \citet{wen2026fantastic}. Full architecture details are in Appendix~\ref{app:model_architecture}, and hyperparameter sweeps are detailed in Appendix~\ref{app:hyperparameters}.

\paragraph{Optimizers} We focus on comparing two different optimization algorithms: AdamW and Muon. We adopt the decoupled weight decay version of AdamW with gradient norm clipping, and we detail the algorithm in Algorithm~\ref{alg:adamw}. \citet{wen2026fantastic} performed extensive hyperparameter search to find the best hyperparameters for pre-training, and we adopt these for our $1\times$ and $8\times$ Chinchilla models. Our Muon algorithm is presented in Algorithm~\ref{alg:muon} and uses the 0.2 RMS update scaling factor from \citet{liu2025muon}, and the same base learning rate as our AdamW pretraining.

Recent work has explored interleaving updates from different optimizers during pretraining \citep{pang2026htmuon}. Following this approach, we construct mixed AdamW-Muon optimizers by independently selecting a Muon update at each step with probability $p \in [0.0, 0.1, 0.25, 0.5, 0.75, 0.9, 1.0]$, and an AdamW update otherwise. Since we use $\beta_1 = 0.9$ for our AdamW training, the expected update RMS according to \citet{su2025adamrms} is $\sqrt{(1-\beta_1)/(1+\beta_1)} \approx 0.22$.
We use this to unify update scales between optimizers, admitting a common base learning rate within and across mixture settings.

\paragraph{Pretraining} Using the same data mixture as \citet{wen2026fantastic}, we pre-train our 1.2B-parameter Llama3 models at $1\times$ and $8\times$ Chinchilla scales \citep{hoffmann2022training}, using approximately 24B and 193B tokens respectively.

\paragraph{Supervised Fine-tuning} We use the Tulu-3 SFT mixture \citep{tulu3} and train for two epochs. We select the best learning rate based on validation loss on a held-out set, and perform final validation on MMLU \citep{mmlu}, BBH \citep{bbh}, GSM8K \citep{gsm8k}, MATH \citep{math}, HumanEval \citep{humaneval}, and MBPP \citep{mbpp}. We pack the SFT sequences to 4096 tokens for our 1.2B models and Llama-2-7B, and pack sequences to 8192 tokens for Qwen models. 

\paragraph{Reinforcement Learning} We train on 35,000 randomly sampled examples from NuminaMath-CoT \citep{numina_math_datasets} for two epochs, using batches of 256 prompts $\times$ 4 rollouts. We select the best learning rate based on final validation reward of a 4,500-example subset of the MATH dataset \citep{lightman2024lets}. We report final benchmark performance on MATH-500 and OlympiadBench \citep{he-etal-2024-olympiadbench}. %

\begin{figure}
  \centering
  \begin{minipage}[t]{0.49\linewidth}
    \centering
    \includegraphics[width=\linewidth]{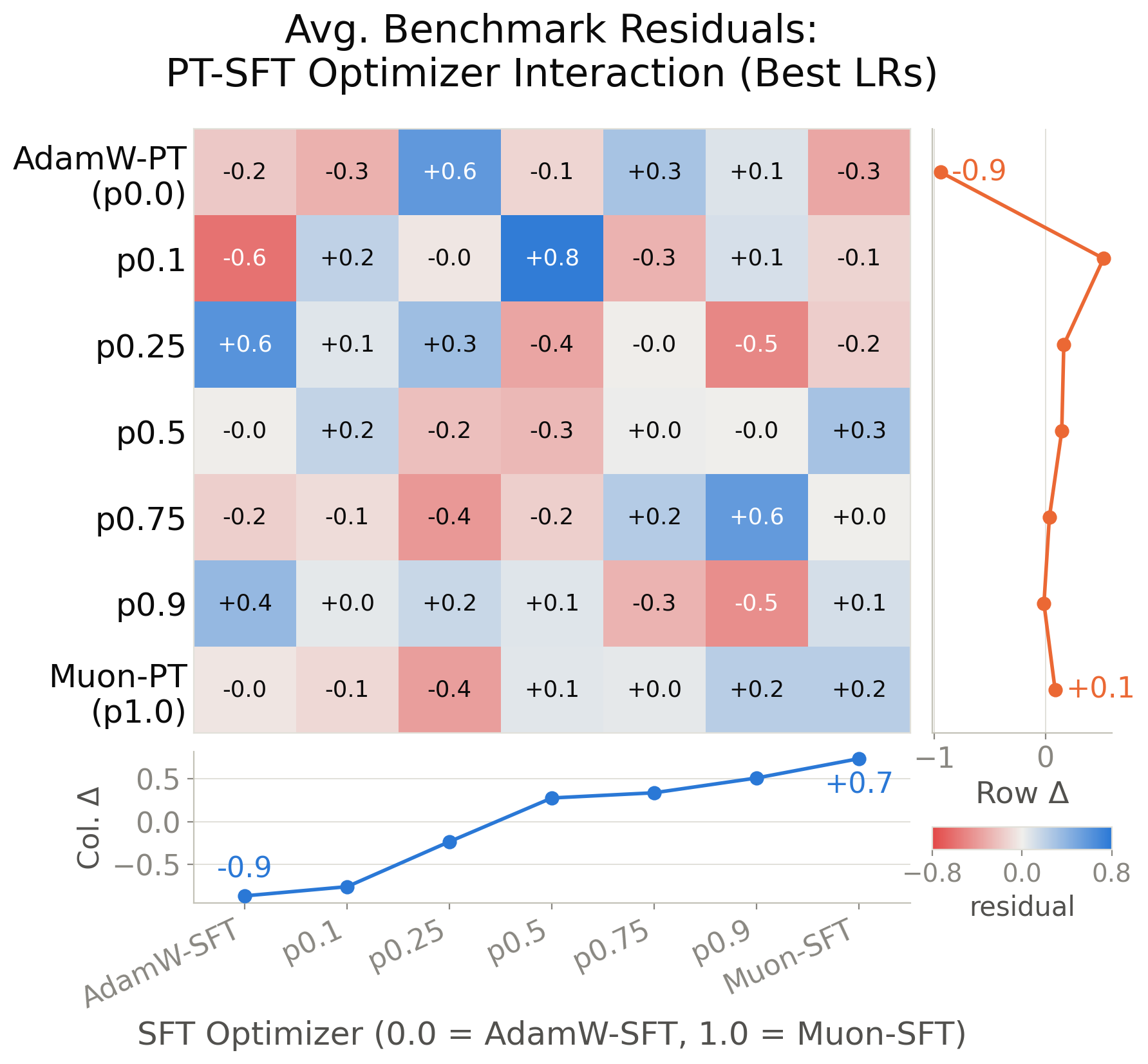}
  \end{minipage}\hfill
  \begin{minipage}[t]{0.5\linewidth}
    \centering
    \includegraphics[width=\linewidth]{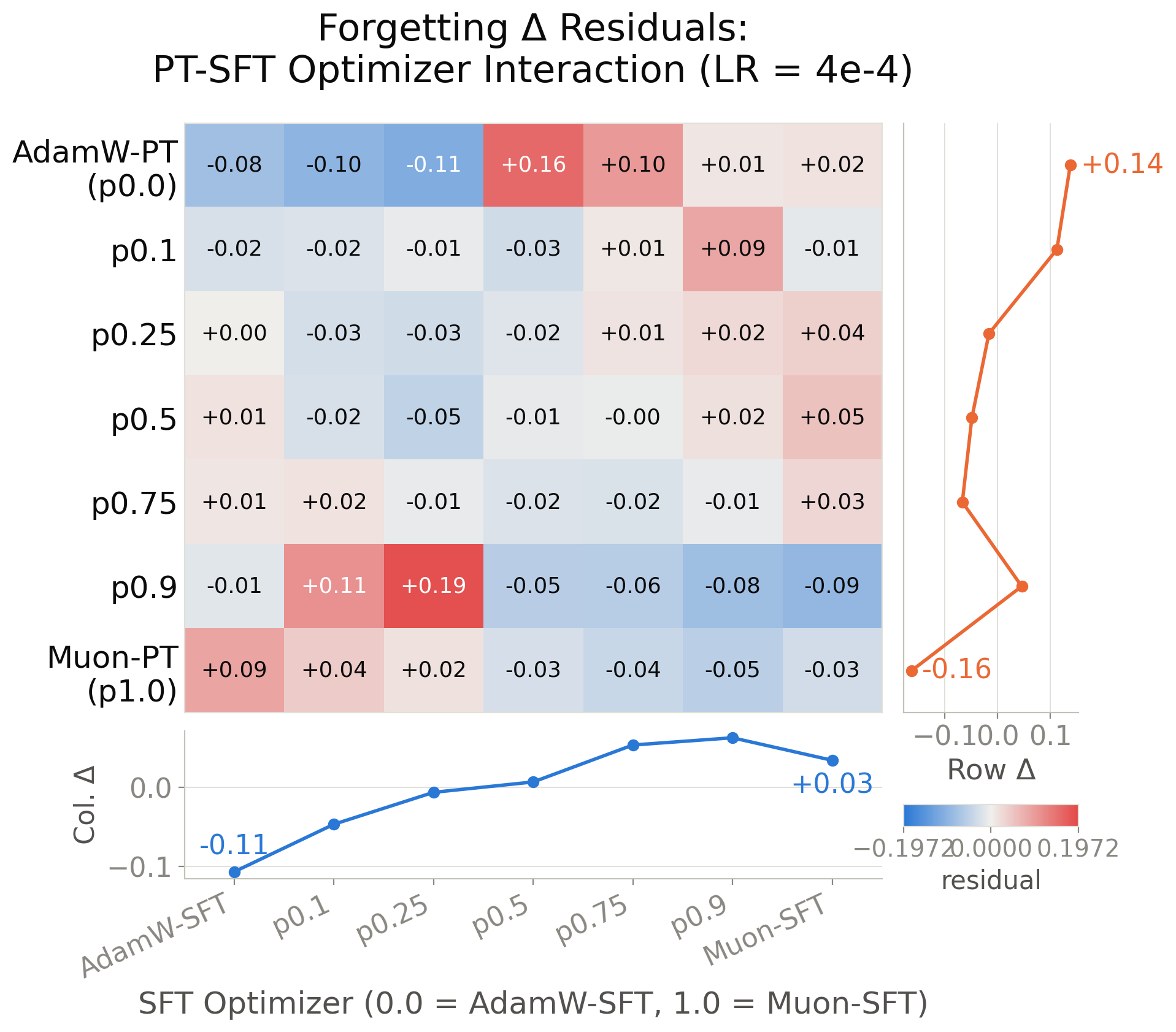}
  \end{minipage}
    \caption{\textbf{SFT benchmark performance and forgetting residuals.} Rows index the proportion of Muon updates for 1.2B Llama-3 models during pretraining ($0.0=\mathrm{AdamW}$, $1.0=\mathrm{Muon}$); columns index the fraction during SFT.
    Average of three independent SFT runs per cell.
    \textbf{Left:} Average of six downstream benchmarks (higher is better), using the best learning rate (Table~\ref{tab:llama3_1b_sft_lr_grid}) for each setting. Optimizer interaction effects are weak and inconsistent. 
    \textbf{Right:} change in pretraining validation loss after SFT (lower is better), at learning rate $4\times10^{-4}$, showing a diagonal matched-optimizer preference pattern.
    Heatmap cells show the residual after accounting for row/column marginal averages, shown along the sides of each heatmap. Raw benchmark averages can be found in Table~\ref{tab:sft_grid}.}
  \label{fig:residual_heatmap}
\end{figure}

\section{Cost of Optimizer Mismatch }
\label{sec:cost_of_optimizer_mismatch}

To isolate the cost of switching optimizers across training stages, we relax the AdamW/Muon optimizer assignment from a binary choice to a continuous degree of mismatch. Specifically, we treat AdamW and Muon as defining two distinct update distributions, and construct intermediate optimizers by randomly selecting an AdamW/Muon update at each step.
This allows us to isolate and quantify switching cost in greater detail, by smoothly annealing the degree of mismatch across stages.

We implement this by pre-training our 1.2B models at 1$\times$ Chinchilla while selecting a Muon update at each step with probability $p \in [0.0, 0.1, 0.25, 0.5, 0.75, 0.9, 1.0]$, where $p=0.0$ corresponds to full-AdamW and $p=1.0$ corresponds to full-Muon. Then, during the SFT stage, we independently vary the optimizer mixture over the same values of $p$. We measure average performance on the six benchmarks across three random seeds for each PT-SFT optimizer pairing.

\paragraph{Isolating the optimizer mismatch effect} To separate optimizer mismatch from the independent effects of the optimizer in either stage, we decompose the score grid into row effects, column effects, and their interaction. For global mean $\mu$, the performance for each cell $X_{ij}$ decomposes as
\begin{equation}
X_{ij} = \mu + \underbrace{\alpha_i}_\text{PT optimizer effect} + \underbrace{\beta_j}_\text{SFT optimizer effect} + \underbrace{\gamma_{ij}}_\text{PT-SFT interaction}
\end{equation}
where $\alpha_i$ and $\beta_j$ are the centered row and column marginal effects respectively, and $\gamma_{ij}$ is the remaining unexplained quantity:
\begin{equation}
    \alpha_i = \bar{X}_{i,:} - \mu, \qquad \beta_j = \bar{X}_{:, j} - \mu,
    \qquad \gamma_{ij} = X_{ij} - \bar{X}_{i,:} - \bar{X}_{:,j} + \mu
\end{equation}

By examining the trends in $\gamma_{ij}$ across the score grid, we can identify interaction effects between optimizers, independent from the performance of each optimizer in general.

\paragraph{Results} If matching optimizers across stages provides a performance benefit, then we would expect better residuals along the diagonal $p_\text{PT} = p_\text{SFT}$, particularly at the fully matched AdamW/AdamW and Muon/Muon corners, and unfavorable residuals as the degree of mismatch increases.
Figure~\ref{fig:residual_heatmap} (left) displays the observed residuals in each heatmap cell: contrary to the optimizer-matching hypothesis, interaction effects between the PT and SFT optimizers are mild, with no consistent overall pattern. The matched-optimizer cells are not meaningfully better than predicted from the independent row- and column-wise effects: AdamW-PT/AdamW-SFT has a negative residual of $-0.2$ while Muon-PT/Muon-SFT has positive residual $+0.2$, with other diagonal interactions ranging from $-0.6$ to $+0.3$. These interactions are small relative to the SFT-optimizer main effect ($-0.9$ to $+0.7$), and are generally comparable to seed noise in the underlying benchmark averages (Table~\ref{tab:sft_grid}). Thus, we find little evidence that matching the optimizer across stages provides meaningful performance benefit.

Final performance is instead dominated by the independent effects of the optimizer used in each stage. The marginal SFT optimizer effect increases monotonically with the proportion of Muon updates, indicating that Muon during SFT generally improves downstream performance regardless of pretraining optimizer. Models pretrained with at least some Muon updates also perform better on average across SFT optimizers, although this primarily appears as a pretraining artifact rather than increased compatibility with Muon SFT; these models already achieve lower pretraining validation loss than AdamW-PT (Table~\ref{tab:llama3_1b_c4en_loss}). %

\paragraph{Effect of SFT distribution} The shift in training data introduced by SFT could potentially obscure a true underlying cost of switching optimizers across training stages. To confirm the lack of broad interaction trends in Figure~\ref{fig:residual_heatmap} (left), we instead continue training on the original pre-training data distribution. The results of this continued pretraining stage are shown in Figure~\ref{fig:residual_heatmap_cpt}. The marginal averages show that final performance improves as more Muon steps are taken, for both pre-training and post-training. When using the same learning rate of 5e-5 across settings, matched CPT optimizers do show the expected favorable interactions (left), reflected by negative residuals, but these effects disappear once the learning rate is tuned for all settings (right).\footnote{For SFT, we do not observe significant interaction effects at either a fixed LR (Figure~\ref{fig:residual_heatmap_cpt_sft_uniform_lr}) or at the best LR (Figure 1, left)} %
This suggests that optimizer-switching costs, when they appear, may stem from hyperparameter mismatch, rather than the optimizers \textit{per se}.

\begin{figure}
  \centering
  \begin{minipage}[t]{0.5\linewidth}
    \centering
    \includegraphics[width=\linewidth]{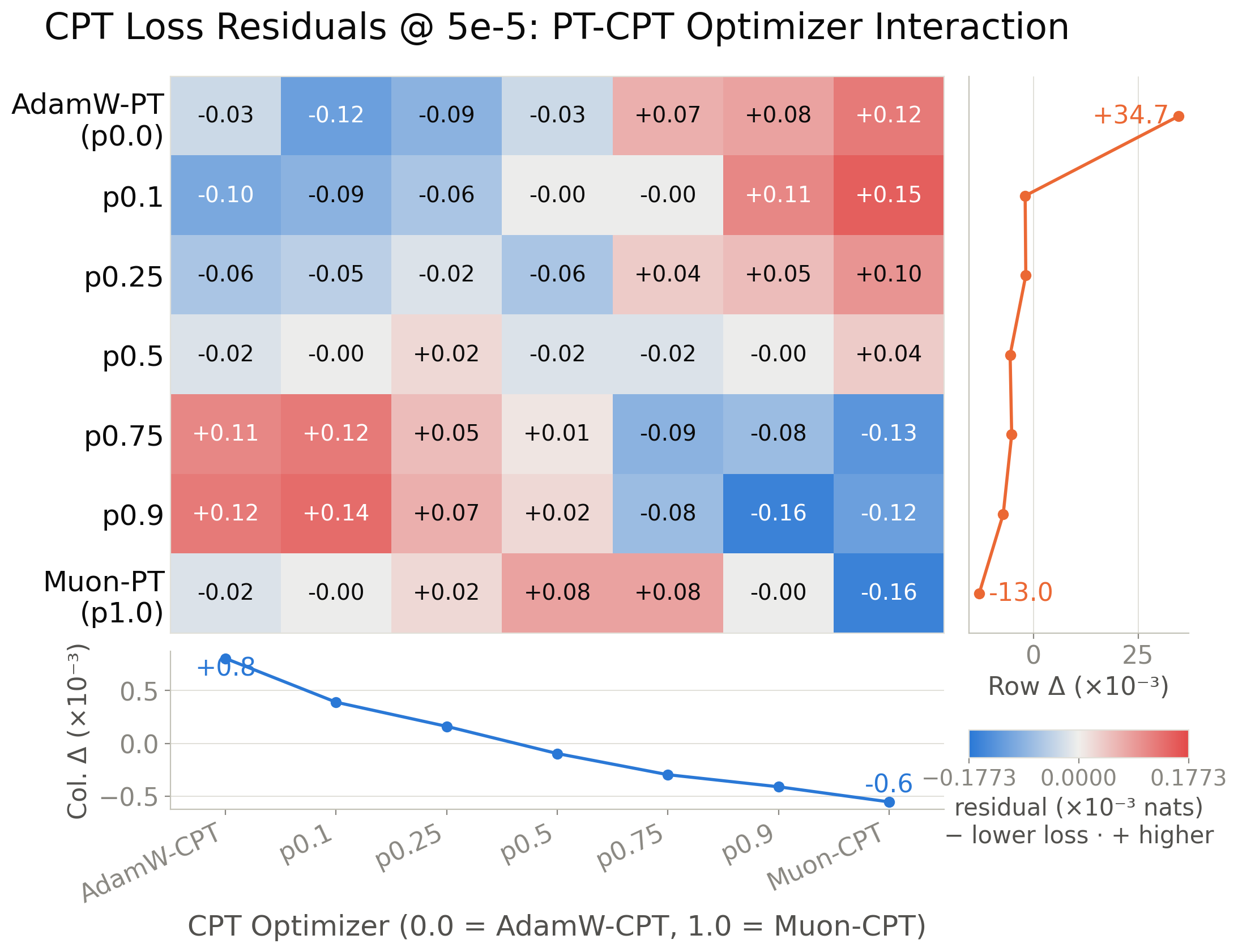}
  \end{minipage}\hfill
  \begin{minipage}[t]{0.5\linewidth}
    \centering
    \includegraphics[width=\linewidth]{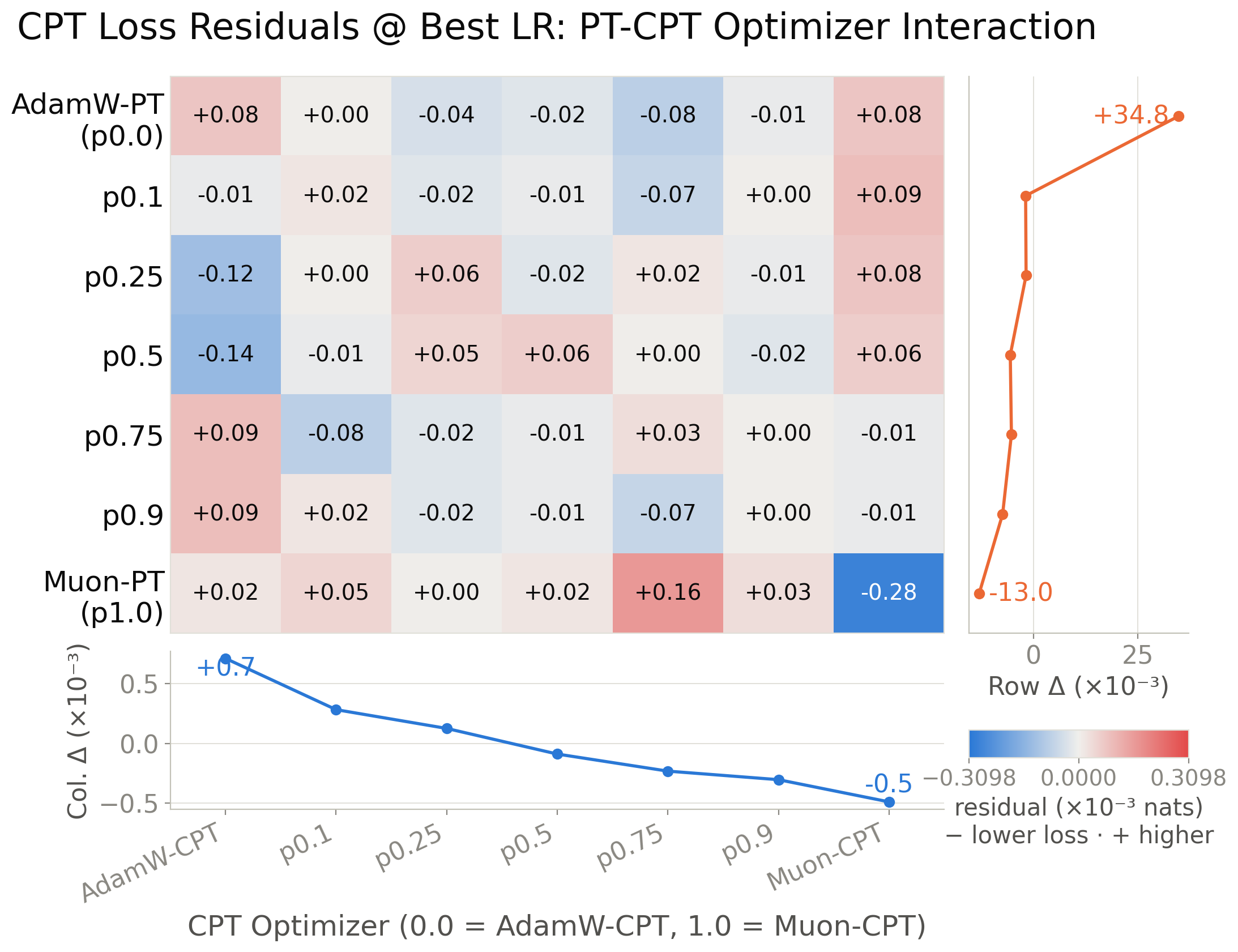}
  \end{minipage}
    \caption{\textbf{Continued pretraining (CPT) performance residuals.} 
    Formatting follows Fig.~\ref{fig:residual_heatmap}: rows index the proportion of Muon updates during pretraining; columns index the fraction for CPT.
    \textbf{Left:} validation loss with a fixed learning rate of $5\times10^{-5}$ for all settings. \textbf{Right:} validation loss using the best learning rate (Table~\ref{tab:llama3_1b_cpt_lr_grid}) for each setting. 
    Matched-optimizer preference vanishes once learning rate is accounted for.
    }

  \label{fig:residual_heatmap_cpt}
\end{figure}

\paragraph{Matched optimizer mitigates forgetting} We next examine whether optimizer consistency affects forgetting of the pretraining distribution, despite limited interaction effect on final downstream performance. We quantify forgetting as the increase in held-out validation loss after SFT, shown in Figure~\ref{fig:residual_heatmap} (right). The column-wise averages show that using more Muon updates during SFT generally results in greater forgetting, while the row-wise averages suggest that using Muon during pretraining mitigates SFT forgetting. Beyond these effects, the residuals show that matched optimizers further reduce forgetting relative to mismatched settings. Thus optimizer switching costs may also manifest as forgetting, rather than as a straightforward reduction in learning capability.

\section{Does model scale change the preferred SFT optimizer?}

\begin{table*}[t]
\centering
\caption{\textbf{SFT benchmark performance for our 1.2B pretrained models at $1\times$ and $8\times$ Chinchilla data ratios.} %
Mean and standard deviation are reported over three SFT seeds. 
}
\label{tab:llama3_1b_benchmark_results}
\scriptsize
\setlength{\tabcolsep}{2pt}
\begin{tabular}{@{}lcccccccc@{}}
\toprule
\textbf{Benchmark}
& \multicolumn{4}{c}{\textbf{1$\times$ Chinchilla}}
& \multicolumn{4}{c}{\textbf{8$\times$ Chinchilla}} \\
\cmidrule(lr){2-5}
\cmidrule(lr){6-9}
& \multicolumn{2}{c}{\textbf{AdamW-PT}}
& \multicolumn{2}{c}{\textbf{Muon-PT}}
& \multicolumn{2}{c}{\textbf{AdamW-PT}}
& \multicolumn{2}{c}{\textbf{Muon-PT}} \\
\cmidrule(lr){2-3}
\cmidrule(lr){4-5}
\cmidrule(lr){6-7}
\cmidrule(lr){8-9}
& \textbf{AdamW} & \textbf{Muon}
& \textbf{AdamW} & \textbf{Muon}
& \textbf{AdamW} & \textbf{Muon}
& \textbf{AdamW} & \textbf{Muon} \\
\midrule

GSM8K (5-shot EM)
& $16.7 \pm 1.0$
& $\mathbf{20.7 \pm 0.8}$
& $17.7 \pm 1.3$
& $\mathbf{20.1 \pm 0.9}$
& $\mathbf{27.8 \pm 1.7}$
& $26.4 \pm 1.1$
& $30.8 \pm 1.0$
& $\mathbf{33.8 \pm 1.8}$ \\

HumanEval (Pass@1)
& $12.4 \pm 0.4$
& $\mathbf{15.7 \pm 1.4}$
& $15.9 \pm 1.2$
& $\mathbf{18.7 \pm 2.0}$
& $\mathbf{18.3 \pm 2.8}$
& $17.5 \pm 2.0$
& $20.7 \pm 1.2$
& $\mathbf{23.4 \pm 2.0}$ \\

MBPP (Pass@1)
& $16.3 \pm 0.7$
& $\mathbf{18.2 \pm 0.8}$
& $18.3 \pm 1.4$
& $\mathbf{21.4 \pm 2.1}$
& $27.0 \pm 0.6$
& $\mathbf{28.9 \pm 0.8}$
& $33.3 \pm 2.6$
& $\mathbf{36.2 \pm 2.1}$ \\

MMLU (0-shot CoT)
& $\mathbf{23.8 \pm 0.5}$
& $22.7 \pm 3.1$
& $22.4 \pm 0.6$
& $\mathbf{24.4 \pm 0.4}$
& $27.5 \pm 1.0$
& $\mathbf{27.6 \pm 1.0}$
& $29.3 \pm 1.2$
& $\mathbf{32.4 \pm 0.3}$ \\

MATH (4-shot CoT)
& $2.4 \pm 0.6$
& $\mathbf{3.3 \pm 0.2}$
& $2.7 \pm 0.3$
& $\mathbf{3.7 \pm 0.7}$
& $\mathbf{4.3 \pm 0.2}$
& $3.3 \pm 0.3$
& $5.1 \pm 0.3$
& $\mathbf{5.6 \pm 0.3}$ \\

BBH (3-shot CoT)
& $\mathbf{23.1 \pm 0.9}$
& $22.9 \pm 1.2$
& $\mathbf{24.7 \pm 0.8}$
& $\mathbf{24.7 \pm 1.9}$
& $27.4 \pm 0.5$
& $\mathbf{27.7 \pm 0.1}$
& $28.9 \pm 2.5$
& $\mathbf{34.4 \pm 0.2}$ \\

\midrule
\textbf{Average}
& $15.8 \pm 0.4$
& $\mathbf{17.2 \pm 0.6}$
& $17.0 \pm 0.3$
& $\mathbf{18.8 \pm 0.2}$
& $\mathbf{22.0 \pm 0.4}$
& $21.9 \pm 0.4$
& $24.7 \pm 0.3$
& $\mathbf{27.6 \pm 0.2}$ \\

\bottomrule
\end{tabular}
\end{table*}

Prior work has reported that for large open-weight LLMs, using the matched optimizer for task-specific SFT may have a slight edge \citep{qu2026muon, liu2026optimizer}, while comparisons for task-general SFT have been limited to a few models, and generally inconclusive \citep{liu2025muon}. In Section~\ref{sec:cost_of_optimizer_mismatch}, our experiments at the 1.2B 1$\times$ Chinchilla scale suggest that Muon may have the edge in task-general SFT regardless of the pretraining optimizer. In this section, we examine whether this phenomenon holds at scale, by increasing the pretraining tokens for our 1.2B model, %
and by fine-tuning open-weight models from various model families up to 14B parameters. %
\paragraph{Data scaling}
\label{sec:data_scaling}

We compare optimizer settings for our 1.2B models at $8\times$ Chinchilla data ratio in Table~\ref{tab:llama3_1b_benchmark_results}. While Muon is clearly the better SFT optimizer for our AdamW-pretrained models at $1\times$, there is a negligible performance gap at $8\times$. On the other hand, Muon clearly remains the better SFT optimizer for Muon-pretrained models at both $1\times$ and $8\times$ Chinchilla data ratios.

\begin{table}[t]
  \centering
  \caption{\textbf{AdamW vs. Muon SFT results for Llama2-7B, Qwen2.5-7B, and Qwen3 family.} Mean and stdev. reported over three trials. Muon achieves higher benchmark average in five of six settings.}
  \label{tab:qwen25_llama2}
  \begin{subtable}[t]{0.49\linewidth}
    \centering
    \caption{Llama2-7B}
    \label{tab:llama2}
    \scriptsize
\setlength{\tabcolsep}{4pt}
\begin{tabular}{@{}lcc@{}}
\toprule
\textbf{Benchmark}
& \multicolumn{2}{c}{\textbf{Llama2-7B}} \\
\cmidrule(lr){2-3}
& \textbf{AdamW} & \textbf{Muon} \\
\midrule
MMLU
& $43.17 \pm 2.86$ & $\mathbf{47.64 \pm 1.35}$ \\
HumanEval
& $36.79 \pm 4.06$ & $\mathbf{42.07 \pm 1.61}$ \\
MBPP
& $44.49 \pm 0.90$ & $\mathbf{48.12 \pm 1.19}$ \\
GSM8K
& $50.37 \pm 1.03$ & $\mathbf{52.26 \pm 0.96}$ \\
MATH
& $11.19 \pm 0.05$ & $\mathbf{11.76 \pm 0.45}$ \\
BBH
& $43.26 \pm 0.80$ & $\mathbf{45.29 \pm 0.64}$ \\
\midrule
\textbf{Average}
& $38.21 \pm 1.36$ & $\mathbf{41.19 \pm 0.33}$ \\
\bottomrule
\end{tabular}

  \end{subtable}
  \hfill
  \begin{subtable}[t]{0.49\linewidth}
    \centering
    \caption{Qwen2.5-7B}
    \label{tab:qwen25}
    \scriptsize
\setlength{\tabcolsep}{4pt}
\begin{tabular}{@{}lcc@{}}
\toprule
\textbf{Benchmark}
& \multicolumn{2}{c}{\textbf{Qwen2.5-7B}} \\
\cmidrule(lr){2-3}
& \textbf{AdamW} & \textbf{Muon} \\
\midrule
MMLU
& $70.32 \pm 0.64$ & $\mathbf{70.65 \pm 0.26}$ \\
HumanEval
& $76.63 \pm 2.88$ & $\mathbf{78.46 \pm 1.41}$ \\
MBPP
& $72.37 \pm 2.06$ & $\mathbf{73.15 \pm 1.78}$ \\
GSM8K
& $\mathbf{78.72 \pm 0.31}$ & $78.37 \pm 0.36$ \\
MATH
& $43.36 \pm 0.59$ & $\mathbf{44.11 \pm 0.83}$ \\
BBH
& $65.15 \pm 0.61$ & $\mathbf{65.81 \pm 0.71}$ \\
\midrule
\textbf{Average}
& $67.76 \pm 0.90$ & $\mathbf{68.42 \pm 0.41}$ \\
\bottomrule
\end{tabular}

  \end{subtable}\\
  \begin{subtable}[t]{\linewidth}
    \centering
    \caption{Qwen3 family at 1.7B, 4B, 8B, and 14B model sizes.}
    
\scriptsize
\setlength{\tabcolsep}{2pt}
\resizebox{\textwidth}{!}{%
\begin{tabular}{@{}lcccccccc@{}}
\toprule
\textbf{Benchmark}
& \multicolumn{2}{c}{\textbf{Qwen3-1.7B}}
& \multicolumn{2}{c}{\textbf{Qwen3-4B}}
& \multicolumn{2}{c}{\textbf{Qwen3-8B}}
& \multicolumn{2}{c}{\textbf{Qwen3-14B}} \\
\cmidrule(lr){2-3}
\cmidrule(lr){4-5}
\cmidrule(lr){6-7}
\cmidrule(lr){8-9}
& \textbf{AdamW} & \textbf{Muon}
& \textbf{AdamW} & \textbf{Muon}
& \textbf{AdamW} & \textbf{Muon}
& \textbf{AdamW} & \textbf{Muon} \\
\midrule

MMLU
& $57.79 \pm 0.98$ & $\mathbf{59.14 \pm 0.41}$
& $70.10 \pm 0.36$ & $\mathbf{70.71 \pm 0.25}$
& $73.41 \pm 0.75$ & $\mathbf{74.38 \pm 0.57}$
& $78.52 \pm 0.11$ & $\mathbf{78.79 \pm 0.11}$ \\

HumanEval
& $67.89 \pm 1.86$ & $\mathbf{69.72 \pm 1.86}$
& $80.08 \pm 1.86$ & $\mathbf{80.69 \pm 0.35}$
& $\mathbf{81.30 \pm 2.88}$ & $78.25 \pm 3.36$
& $86.18 \pm 0.35$ & $\mathbf{86.38 \pm 0.35}$ \\

MBPP
& $\mathbf{66.67 \pm 0.90}$ & $65.76 \pm 1.40$
& $\mathbf{76.52 \pm 0.45}$ & $76.13 \pm 1.62$
& $\mathbf{82.10 \pm 1.78}$ & $81.71 \pm 1.03$
& $83.27 \pm 0.39$ & $\mathbf{83.53 \pm 0.59}$ \\

GSM8K
& $67.50 \pm 1.16$ & $\mathbf{67.93 \pm 0.75}$
& $80.14 \pm 0.60$ & $\mathbf{82.08 \pm 0.89}$
& $\mathbf{84.23 \pm 0.55}$ & $83.62 \pm 0.39$
& $87.39 \pm 0.63$ & $\mathbf{88.12 \pm 0.57}$ \\

MATH
& $36.53 \pm 0.39$ & $\mathbf{38.56 \pm 0.23}$
& $47.62 \pm 0.02$ & $\mathbf{48.39 \pm 0.46}$
& $\mathbf{50.68 \pm 0.13}$ & $50.60 \pm 0.30$
& $\mathbf{56.45 \pm 0.65}$ & $55.09 \pm 1.63$ \\

BBH
& $52.63 \pm 0.72$ & $\mathbf{52.89 \pm 0.30}$
& $\mathbf{69.55 \pm 0.53}$ & $69.44 \pm 0.41$
& $73.11 \pm 0.99$ & $\mathbf{73.68 \pm 0.42}$
& $78.24 \pm 0.41$ & $\mathbf{78.46 \pm 0.11}$ \\

\midrule
\textbf{Average}
& $58.17 \pm 0.23$ & $\mathbf{59.00 \pm 0.13}$
& $70.67 \pm 0.39$ & $\mathbf{71.24 \pm 0.30}$
& $\mathbf{74.14 \pm 0.33}$ & $73.71 \pm 0.73$
& $78.34 \pm 0.02$ & $\mathbf{78.40 \pm 0.24}$ \\

\bottomrule
\end{tabular}%
}

  \end{subtable}
\end{table}

\paragraph{Open weight models.}
\label{sec:open_weight_models}
We compare AdamW and Muon SFT across three seeds for AdamW-pretrained Llama2-7B, Qwen2.5-7B, and Qwen3 models up to 14B (Table~2). Muon achieves higher benchmark averages in five of six settings, although Qwen differences are small relative to seed variation. Muon's clearer advantage on Llama2-7B may reflect less extensive pretraining, matching trends on the 1.2B models.

\section{Does optimizer mismatch matter in reinforcement learning?}

\begin{table*}[t]
\centering
\caption{\textbf{Math performance and forgetting after GRPO.}
We report math benchmark accuracy (\%) under greedy decoding with a 3K token budget, changes in pretraining and SFT validation losses (1.2B only), and six-task out-of-domain performance. Results are mean $\pm$ standard deviation over three seeds at each optimizer's best learning rate.
}
\label{tab:grpo_math_benchmarks_with_norl}
\scriptsize
\setlength{\tabcolsep}{3pt}

\begin{subtable}{\textwidth}
\centering
\caption{\textbf{1.2B pretrained models at 8$\times$ Chinchilla.}}
\label{tab:grpo_1b}
\begin{tabular}{@{}lcccccc@{}}
\toprule
\textbf{Benchmark}
& \multicolumn{3}{c}{\textbf{AdamW-PT}}
& \multicolumn{3}{c}{\textbf{Muon-PT}} \\
\cmidrule(lr){2-4}
\cmidrule(lr){5-7}
& \textbf{No RL} & \textbf{AdamW} & \textbf{Muon}
& \textbf{No RL} & \textbf{AdamW} & \textbf{Muon} \\
\midrule

MATH-500
& $4.20$
& $7.60 \pm 0.53$
& $\mathbf{8.00 \pm 1.11}$
& $8.40$
& $\mathbf{11.33 \pm 0.76}$
& $11.27 \pm 0.99$ \\

OlympiadBench
& $1.19$
& $\mathbf{1.98 \pm 0.45}$
& $1.78 \pm 0.15$
& $1.48$
& $2.67 \pm 0.15$
& $\mathbf{2.72 \pm 0.37}$ \\

\midrule
\multicolumn{7}{@{}l}{\textit{Forgetting}} \\

PT val. loss $\Delta$ $(\downarrow)$
& 
& $\mathbf{+0.0082 \pm 0.0024}$
& $+0.0124 \pm 0.0012$
& 
& $\mathbf{+0.0377 \pm 0.0064}$
& $+0.0400 \pm 0.0065$ \\

SFT val. loss $\Delta$ $(\downarrow)$
& 
& $+0.0092 \pm 0.0006$
& $\mathbf{+0.0079 \pm 0.0004}$
& 
& $\mathbf{+0.0242 \pm 0.0043}$
& $+0.0266 \pm 0.0025$ \\ 

6-task avg. $\Delta$ $(\uparrow)$
& 
& $+0.04 \pm 0.26$
& $\mathbf{+0.35 \pm 0.20}$
& 
& $\mathbf{-0.34 \pm 0.30}$
& $-0.99 \pm 0.79$ \\

\bottomrule
\end{tabular}
\end{subtable}

\vspace{0.8em}

\begin{subtable}{\textwidth}
\centering
\caption{\textbf{Qwen3 models.}}
\label{tab:grpo_qwen3}
\begin{tabular}{@{}lcccccc@{}}
\toprule
\textbf{Benchmark}
& \multicolumn{3}{c}{\textbf{Qwen3-1.7B}}
& \multicolumn{3}{c}{\textbf{Qwen3-4B}} \\
\cmidrule(lr){2-4}
\cmidrule(lr){5-7}
& \textbf{No RL} & \textbf{AdamW} & \textbf{Muon}
& \textbf{No RL} & \textbf{AdamW} & \textbf{Muon} \\
\midrule

MATH-500
& $51.20$
& $77.80 \pm 1.59$
& $\mathbf{78.20 \pm 0.60}$
& $51.60$
& $\mathbf{85.20 \pm 0.40}$
& $\mathbf{85.20 \pm 0.35}$ \\

OlympiadBench
& $18.69$
& $40.60 \pm 1.77$
& $\mathbf{43.13 \pm 2.03}$
& $17.95$
& $48.86 \pm 0.95$
& $\mathbf{51.29 \pm 1.20}$ \\

\midrule
\multicolumn{7}{@{}l}{\textit{Forgetting}} \\

6-task avg. $\Delta$ $(\uparrow)$
&
& $\mathbf{+0.70 \pm 0.24}$
& $+0.14 \pm 0.35$
& 
& $\mathbf{+0.37 \pm 0.09}$
& $+0.35 \pm 0.37$ \\

\bottomrule
\end{tabular}
\end{subtable}

\end{table*}

We evaluate the impact of optimizer mismatch on math RLVR using our 1.2B 8$\times$ Chinchilla models and Qwen3-1.7B/4B (Table~\ref{tab:grpo_math_benchmarks_with_norl}). Across three seeds at each optimizer's best learning rate, AdamW and Muon achieve comparable performance on MATH-500 and OlympiadBench, with differences within seed-to-seed variation. %
Interestingly, while AdamW generally prefers equal or higher learning rates during SFT, Muon prefers $2$--$4\times$ larger learning rates during GRPO. However, at a fixed learning rate, AdamW generally produces greater forward KL from the initial policy. As discussed in Appendix~D, forward KL is more closely associated with forgetting than learning rate or accumulated parameter-change norms.

We measure forgetting using held-out pretraining and SFT validation losses for our 1.2B models, and changes in six out-of-domain benchmarks for all models, following \citet{shenfeld2026rls} (Appendix~\ref{app:rlvr_hyperparameters}). For Qwen3, where the pretraining data is unknown, we report only the change in 6-task average. Across all settings, forgetting remains relatively small at the selected learning rates, and differences between optimizers are generally within seed variation. The optimizer with lower forward KL also achieves better mean retention on the six out-of-domain tasks in all four model settings, as we expand on in Appendix~\ref{app:rlvr}.

\section{Conclusion}

We study the cost of switching between AdamW and Muon across LLM training stages. Across controlled experiments interpolating between AdamW and Muon updates, we find little evidence of a meaningful optimizer mismatch penalty: final performance is largely determined by the independent marginal effects of the pretraining and post-training optimizers, rather than by whether they match. Optimizer consistency can reduce forgetting of the pretraining distribution during SFT, but this does not translate to improved final performance under task-general SFT. Instead, at the 1.2B scale, increasing the proportion of Muon updates during SFT generally improves downstream performance regardless of pretraining optimizer. Experiments on open-weight models across the Llama and Qwen families up to 14B further show that Muon remains competitive with AdamW for SFT of AdamW-pretrained models. 
In math RLVR, both optimizers achieve comparable performance and limited forgetting despite different optimal learning rates and parameter-change geometries. Forgetting is more closely associated with forward KL from the initial policy than optimizer choice. Overall, our results suggest that post-training optimizer selection can be made largely independently of the pretraining optimizer.

\bibliographystyle{plainnat}
\bibliography{references}

\appendix

\section{Experimental Details}

\subsection{Model Architecture}
\label{app:model_architecture}

For pretraining, we use the Llama-3 architecture and tokenizer. Following the 1.2B configuration of ~\cite{wen2026fantastic}, our models have 32 transformer layers, a hidden dimension of 1536, an intermediate dimension of 6144, and 24 attention heads. We use full multi-head attention with 24 key-value heads. The context length is 4096 tokens and the vocabulary size is 128,256. We use SwiGLU activations and rotary positional embeddings (RoPE) with $\theta=500{,}000$. The input embedding and language-model head are untied.

This configuration contains approximately 1.60B parameters in total, of which approximately 1.21B are non-embedding parameters. Following~\cite{wen2026fantastic}, we refer to the model by its non-embedding parameter count as a 1.2B model. During training, model parameters are stored in FP32 and activations are computed in BF16.

\subsection{Optimizers}
\label{app:optimizers}

\begin{algorithm}[h]
\caption{AdamW}
\label{alg:adamw}
\begin{algorithmic}[1]
\Require Gradient $G_t$, parameters $W_{t-1}$, first moment $m_{t-1}$,
second moment $v_{t-1}$, learning rate $\eta$, weight decay $\lambda$,
coefficients $\beta_1,\beta_2$, constant $\epsilon$
\State $m_t \gets \beta_1 m_{t-1} + (1-\beta_1)G_t$
\State $\hat{m}_t \gets m_t / (1-\beta_1^t)$
\State $v_t \gets \beta_2 v_{t-1} + (1-\beta_2)G_t^{\odot 2}$
\State $\hat{v}_t \gets v_t / (1-\beta_2^t)$
\State $U_t \gets \hat{m}_t \oslash \left(\sqrt{\hat{v}_t}+\epsilon\right)$
\State $W_t \gets (1-\eta\lambda)W_{t-1} - \eta U_t$
\State \Return $W_t, m_t, v_t$
\end{algorithmic}
\end{algorithm}

\begin{algorithm}[h]
\caption{Muon}
\label{alg:muon}
\begin{algorithmic}[1]
\Require Gradient $G_t \in \mathbb{R}^{m\times n}$, parameters $W_{t-1}$,
momentum $m_{t-1}$, learning rate $\eta$, weight decay $\lambda$,
momentum coefficient $\mu$
\State $m_t \gets \mu m_{t-1} + G_t$
\State $B_t \gets G_t + \mu m_t$
\State $O_t \gets \mathrm{NS}_5(B_t)$
\State $U_t \gets 0.2\sqrt{\max(m,n)}\,O_t$
\State $W_t \gets (1-\eta\lambda)W_{t-1} - \eta U_t$
\State \Return $W_t, m_t$
\end{algorithmic}
\end{algorithm}

Following \citet{liu2025muon} and \citet{wen2026fantastic}, we perform orthogonalization on the concatenated attention head matrices.

\subsection{Hyperparameters}
\label{app:hyperparameters}
\subsubsection{Pretraining}
\label{app:pretraining_hyperparameters}

We pretrain our 1.2B models using the same data mixture as~\cite{wen2026fantastic}, consisting of DCLM-baseline, StarCoderData, and ProofPile-2. All data are tokenized using the Llama-3 tokenizer. We train at both $1\times$ and $8\times$ Chinchilla data ratios, corresponding to approximately 24.16B and 193.29B tokens, respectively. With a global batch size of 256 sequences and a sequence length of 4096, each optimization step contains 1,048,576 tokens. We therefore train for 23,042 steps at $1\times$ Chinchilla and 184,336 steps at $8\times$ Chinchilla.

We use a linear warmup followed by linear learning-rate decay to zero. For the $1\times$ Chinchilla runs, AdamW uses a peak learning rate of $2\times10^{-3}$, $\beta_1=0.9$, $\beta_2=0.98$, $\epsilon=10^{-10}$, weight decay $0.2$, gradient-norm clipping at $2.0$, and 2,000 warmup steps. Muon uses a peak matrix learning rate of $2\times10^{-3}$, momentum $\mu=0.98$, weight decay $0.1$, five Newton--Schulz iterations, and no learning-rate warmup. Parameters not optimized by Muon are optimized with AdamW using a learning rate of $1.2\times10^{-3}$, $(\beta_1,\beta_2)=(0.8,0.98)$, $\epsilon=10^{-15}$, and weight decay $0.1$.

For the $8\times$ Chinchilla runs, we use the budget-specific hyperparameters from~\cite{wen2026fantastic}. AdamW uses a peak learning rate of $2\times10^{-3}$, weight decay $0.2$, gradient-norm clipping at $1.0$, and 1,000 warmup steps. Muon uses a matched peak matrix learning rate of $2\times10^{-3}$, weight decay $0.1$, gradient-norm clipping at $2.0$, and no warmup; its remaining hyperparameters and the AdamW parameter group are unchanged from the $1\times$ setting.

For mixed-optimizer pretraining runs, at each optimization step we independently select a Muon update with probability $p$ and an AdamW update otherwise. We use 2,000 learning rate warmup steps. As described in Section~2, we use a common learning rate and weight decay for the two update branches so that their update magnitudes remain comparable.

Final validation loss on C4-EN for our pretrained models is shown in Table~\ref{tab:llama3_1b_c4en_loss}.

\begin{table}[]
    \centering
    \caption{C4-EN validation loss for our 1.2B Llama3 models.}
    \label{tab:llama3_1b_c4en_loss}
    \small
    \setlength{\tabcolsep}{4pt}
    \begin{tabular}{@{}lcc@{}}
    \toprule
    \textbf{Pretraining Optimizer} & \textbf{1$\times$ Chinchilla} & \textbf{8$\times$ Chinchilla} \\
    \midrule
    \textbf{AdamW}     & 2.9426 & 2.8011 \\
    $p = 0.1$   & 2.9072 & --- \\
    $p = 0.25$  & 2.9075 & --- \\
    $p = 0.5$   & 2.9047 & --- \\
    $p = 0.75$  & 2.9042 & --- \\
    $p = 0.9$   & 2.9020 & --- \\
    \textbf{Muon}   & 2.8947 & 2.7375 \\
    \bottomrule
    \end{tabular}
\end{table}

\subsubsection{SFT}
\label{app:sft_hyperparameters}

We perform supervised fine-tuning using the Tulu-3 SFT mixture~\citep{tulu3} for two epochs for both our 1.2B models and open weight models.

\paragraph{1.2B Pretrained Models}
For our 1.2B models, training examples are greedily packed into sequences of length 4096 without splitting individual conversations across sequence boundaries. Labels corresponding to prompt and user tokens are masked from the loss. For both AdamW and Muon, we use a linear warmup for 3\% of the total SFT steps, followed by linear decay to zero. We sweep the base learning rate separately for each optimizer and experimental setting and select the learning rate with the lowest held-out validation loss. When mixing optimizers, a Muon update is independently selected at each step with probability $p$ and an AdamW update is selected otherwise.

The optimizer hyperparameters largely follow the corresponding pretraining configurations described in Appendix~\ref{app:pretraining_hyperparameters}, with several differences for SFT. For AdamW, we use $(\beta_1,\beta_2)=(0.9,0.98)$, $\epsilon=10^{-10}$, weight decay $0.2$, and global gradient-norm clipping at $1.0$. For Muon, we use Nesterov momentum $\mu=0.98$, weight decay $0.1$, and five Newton--Schulz iterations with $\epsilon_{\mathrm{NS}}=10^{-5}$. As during pretraining, Muon is applied to matrix-valued parameters, while embeddings, the language-model head, and normalization parameters are optimized by a separate AdamW parameter group with $(\beta_1,\beta_2)=(0.8,0.98)$, $\epsilon=10^{-15}$, weight decay $0.1$, and learning rate $5\times10^{-5}$. Global gradient-norm clipping at $1.0$ is also used for Muon runs.

\begin{figure}
    \centering
    \includegraphics[width=1.0\linewidth]{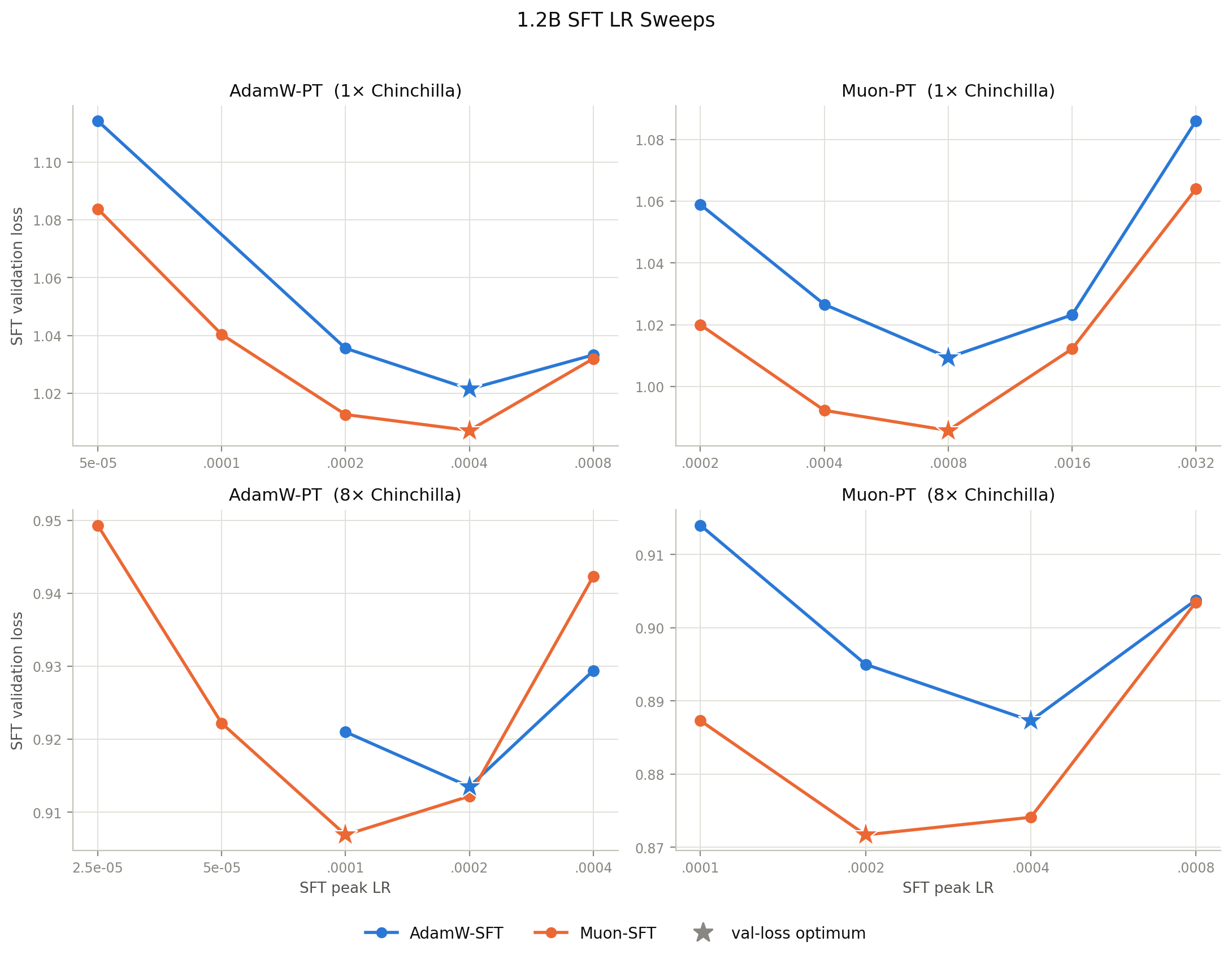}
    \caption{SFT learning rate sweep for AdamW and Muon on 1.2B pretrained Llama3 models.}
    \label{fig:sft_lr_2x2.png}
\end{figure}

\begin{table}[]
    \centering
    \caption{Best SFT learning rate for our 1.2B $1\times$ Chinchilla Llama3 models.}
    \label{tab:llama3_1b_sft_lr_grid}
\small
\setlength{\tabcolsep}{4pt}
\begin{tabular}{@{}lccccccc@{}}
\toprule
\textbf{PT opt $\backslash$ SFT opt}
& \textbf{AdamW}
& \textbf{0.1}
& \textbf{0.25}
& \textbf{0.5}
& \textbf{0.75}
& \textbf{0.9}
& \textbf{Muon} \\
\midrule

\textbf{AdamW-PT}
& $4\mathrm{e}{-4}$
& $4\mathrm{e}{-4}$
& $4\mathrm{e}{-4}$
& $4\mathrm{e}{-4}$
& $4\mathrm{e}{-4}$
& $4\mathrm{e}{-4}$
& $4\mathrm{e}{-4}$ \\

$p0.1$
& $4\mathrm{e}{-4}$
& $4\mathrm{e}{-4}$
& $4\mathrm{e}{-4}$
& $4\mathrm{e}{-4}$
& $4\mathrm{e}{-4}$
& $4\mathrm{e}{-4}$
& $4\mathrm{e}{-4}$ \\

$p0.25$
& $4\mathrm{e}{-4}$
& $4\mathrm{e}{-4}$
& $4\mathrm{e}{-4}$
& $4\mathrm{e}{-4}$
& $4\mathrm{e}{-4}$
& $4\mathrm{e}{-4}$
& $4\mathrm{e}{-4}$ \\

$p0.5$
& $4\mathrm{e}{-4}$
& $4\mathrm{e}{-4}$
& $4\mathrm{e}{-4}$
& $4\mathrm{e}{-4}$
& $4\mathrm{e}{-4}$
& $4\mathrm{e}{-4}$
& $4\mathrm{e}{-4}$ \\

$p0.75$
& $4\mathrm{e}{-4}$
& $4\mathrm{e}{-4}$
& $4\mathrm{e}{-4}$
& $4\mathrm{e}{-4}$
& $4\mathrm{e}{-4}$
& $4\mathrm{e}{-4}$
& $4\mathrm{e}{-4}$ \\

$p0.9$
& $\mathbf{8\mathrm{e}{-4}}$
& $4\mathrm{e}{-4}$
& $4\mathrm{e}{-4}$
& $4\mathrm{e}{-4}$
& $4\mathrm{e}{-4}$
& $4\mathrm{e}{-4}$
& $4\mathrm{e}{-4}$ \\

\textbf{Muon-PT}
& $\mathbf{8\mathrm{e}{-4}}$
& $\mathbf{8\mathrm{e}{-4}}$
& $\mathbf{8\mathrm{e}{-4}}$
& $\mathbf{8\mathrm{e}{-4}}$
& $\mathbf{8\mathrm{e}{-4}}$
& $\mathbf{8\mathrm{e}{-4}}$
& $\mathbf{8\mathrm{e}{-4}}$ \\

\bottomrule
\end{tabular}
\end{table}

\begin{table}[]
    \centering
    \caption{Best CPT learning rate for our 1.2B $1\times$ Chinchilla Llama3 models.}
    \label{tab:llama3_1b_cpt_lr_grid}
\small
\setlength{\tabcolsep}{4pt}
\begin{tabular}{@{}lccccccc@{}}
\toprule
\textbf{PT opt $\backslash$ CPT opt}
& \textbf{AdamW}
& \textbf{0.1}
& \textbf{0.25}
& \textbf{0.5}
& \textbf{0.75}
& \textbf{0.9}
& \textbf{Muon} \\
\midrule

\textbf{AdamW-PT}
& $5\mathrm{e}{-5}$
& $5\mathrm{e}{-5}$
& $5\mathrm{e}{-5}$
& $5\mathrm{e}{-5}$
& $5\mathrm{e}{-5}$
& $5\mathrm{e}{-5}$
& $5\mathrm{e}{-5}$ \\

$p0.1$
& $5\mathrm{e}{-5}$
& $5\mathrm{e}{-5}$
& $5\mathrm{e}{-5}$
& $5\mathrm{e}{-5}$
& $5\mathrm{e}{-5}$
& $5\mathrm{e}{-5}$
& $5\mathrm{e}{-5}$ \\

$p0.25$
& $1\mathrm{e}{-4}$
& $1\mathrm{e}{-4}$
& $5\mathrm{e}{-5}$
& $5\mathrm{e}{-5}$
& $5\mathrm{e}{-5}$
& $5\mathrm{e}{-5}$
& $5\mathrm{e}{-5}$ \\

$p0.5$
& $1\mathrm{e}{-4}$
& $1\mathrm{e}{-4}$
& $1\mathrm{e}{-4}$
& $5\mathrm{e}{-5}$
& $5\mathrm{e}{-5}$
& $5\mathrm{e}{-5}$
& $5\mathrm{e}{-5}$ \\

$p0.75$
& $1\mathrm{e}{-4}$
& $1\mathrm{e}{-4}$
& $1\mathrm{e}{-4}$
& $1\mathrm{e}{-4}$
& $5\mathrm{e}{-5}$
& $5\mathrm{e}{-5}$
& $5\mathrm{e}{-5}$ \\

$p0.9$
& $1\mathrm{e}{-4}$
& $1\mathrm{e}{-4}$
& $1\mathrm{e}{-4}$
& $1\mathrm{e}{-4}$
& $1\mathrm{e}{-4}$
& $5\mathrm{e}{-5}$
& $1\mathrm{e}{-4}$ \\

\textbf{Muon-PT}
& $1\mathrm{e}{-4}$
& $1\mathrm{e}{-4}$
& $1\mathrm{e}{-4}$
& $1\mathrm{e}{-4}$
& $5\mathrm{e}{-5}$
& $5\mathrm{e}{-5}$
& $1\mathrm{e}{-4}$ \\

\bottomrule
\end{tabular}
\end{table}

The results of our SFT learning rate sweeps can be found in Figure~\ref{fig:sft_lr_2x2.png} and Table~\ref{tab:llama3_1b_sft_lr_grid}. For the SFT grid reported in Figure~\ref{fig:residual_heatmap}, optimal learning rates are found by first bracketing the best learning rates for the first and last column 
as well as the matched-optimizer diagonal. %
Since these sweep results show that the optimal learning rate is in $\{4\times 10^{-4}, 8\times 10^{-4}\}$, we only sweep these results for the rest of the table. 

We perform the same procedure for our CPT learning rate sweep and report the best learning rates for each setting in Table~\ref{tab:llama3_1b_cpt_lr_grid}.

Across both SFT and CPT, we observe that the optimal post-training learning rate depends on the pretraining optimizer as well as the post-training optimizer mixture. In particular, models pretrained with a larger proportion of Muon updates tend to prefer a higher learning rate when post-trained with AdamW-heavy mixtures, while increasing the proportion of Muon updates during post-training generally reduces this difference. This dependence motivates selecting the learning rate separately for each PT–post-training optimizer pairing rather than using a single learning rate across the grid.

\paragraph{Open Weight Models}
For open-weight models, we use the same Tulu-3 mixture and assistant-only loss, and greedily pack conversations without splitting them across sequence boundaries. Packed conversations are isolated using a document-level attention mask with position IDs reset at each conversation boundary. We pack sequences to a maximum length of 8192 tokens for Qwen2.5 and Qwen3 models and 4096 tokens for Llama-2, whose context length is limited to 4096 tokens.

For AdamW, we use $(\beta_1,\beta_2)=(0.9,0.999)$, $\epsilon=10^{-8}$, zero weight decay, and global gradient-norm clipping at $1.0$. For Muon, we use the RMS-matched update scaling of \citet{liu2025muon}, $0.2\sqrt{\max(m,n)}$, with Nesterov momentum $\mu=0.95$ and five Newton--Schulz iterations. Muon is applied to two-dimensional parameter matrices, while embeddings, the language-model head, and normalization parameters are optimized with AdamW at the same learning rate. Both AdamW and Muon use the same learning rate within a given model and learning-rate sweep.

\begin{table}[t]
\centering
\caption{SFT learning rate sweep results for open weight models.}
\label{tab:open_weight_sft_lrs}
\small
\setlength{\tabcolsep}{6pt}
\begin{tabular}{@{}lll@{}}
\toprule
\textbf{Model} & \textbf{Swept} & \textbf{Best LR} \\
\midrule
Qwen2.5-7B & $2\mathrm{e}{-5},\, 5\mathrm{e}{-5}$ ($\rightarrow 0$ and $\rightarrow$ peak/10) & $\mathbf{2\mathrm{e}{-5}}$ \\
Qwen3-1.7B   & $1\mathrm{e}{-5},\, 2\mathrm{e}{-5},\, 4\mathrm{e}{-5},\, 8\mathrm{e}{-5}$ & $\mathbf{4\mathrm{e}{-5}}$ \\
Qwen3-4B   & $1\mathrm{e}{-5},\, 2\mathrm{e}{-5},\, 4\mathrm{e}{-5},\, 8\mathrm{e}{-5}$ & $\mathbf{2\mathrm{e}{-5}}$ \\
Qwen3-8B   & $1\mathrm{e}{-5},\, 2\mathrm{e}{-5},\, 4\mathrm{e}{-5}$ & $\mathbf{2\mathrm{e}{-5}}$ \\
Qwen3-14B  & $5\mathrm{e}{-6},\, 1\mathrm{e}{-5},\, 2\mathrm{e}{-5},\, 4\mathrm{e}{-5}$ & $\mathbf{1\mathrm{e}{-5}}$ \\
Llama-2-7B & $1\mathrm{e}{-5},\, 2\mathrm{e}{-5},\, 5\mathrm{e}{-5},\, 1\mathrm{e}{-4}$ & $\mathbf{5\mathrm{e}{-5}}$ \\
\bottomrule
\end{tabular}
\end{table}

We use $3\%$ learning-rate warmup and then follow \citet{liu2025muon} in using a cosine decay schedule to $0.1$ times the peak learning rate. We use a global batch size of 524,288 tokens per optimization step, corresponding to 64 sequences of length 8192 for Qwen models and 128 sequences of length 4096 for Llama-2. We sweep the peak learning rate separately for each model and select the value with the lowest held-out validation loss. Our selected learning rates are shown in Table~\ref{tab:open_weight_sft_lrs}.

\subsection{RLVR}
\label{app:rlvr_hyperparameters}

\begin{figure}
    \centering
    \includegraphics[width=1.0\linewidth]{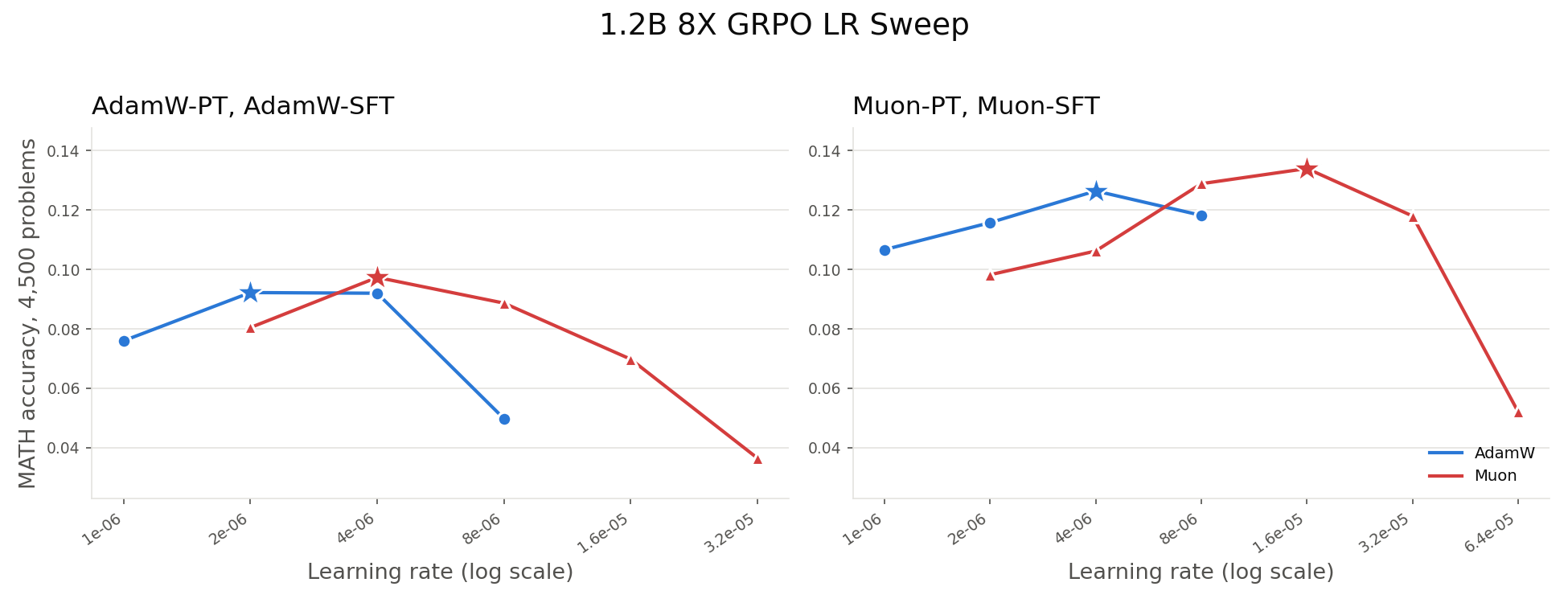}
    \caption{GRPO learning rate sweep for AdamW and Muon on 1.2B pretrained Llama-3 models.}
    \label{fig:grpo_1p2b_lr_sweep}
\end{figure}

\begin{figure}
    \centering
    \includegraphics[width=1.0\linewidth]{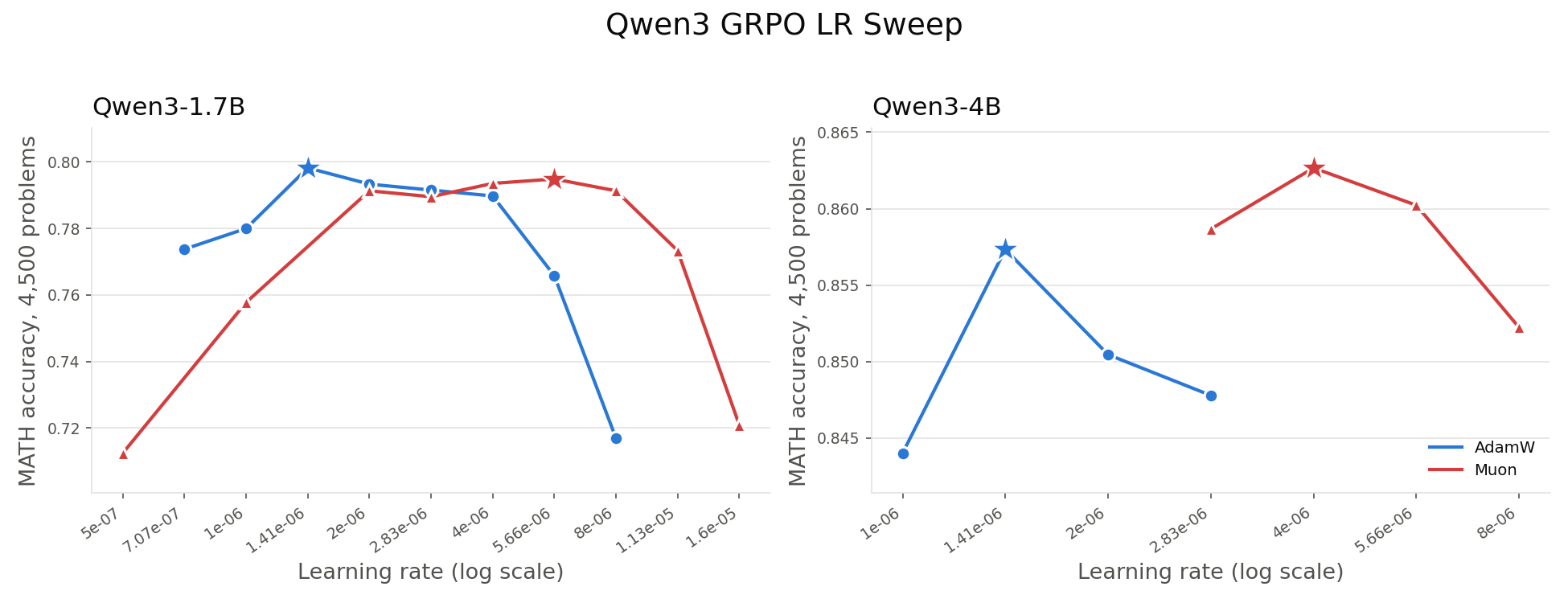}
    \caption{GRPO learning rate sweep for AdamW and Muon on Qwen3-1.7B and 4B.}
    \label{fig:grpo_qwen3_lr_sweep}
\end{figure}

The results of our GRPO learning rate sweeps can be found in Figure~\ref{fig:grpo_1p2b_lr_sweep} and Figure~\ref{fig:grpo_qwen3_lr_sweep}.

\paragraph{Data and prompting.}
We train on 35,000 problems from the \texttt{synthetic\_math} subset of NuminaMath-CoT~\citep{numina_math_datasets}, sampled uniformly at
random. Each prompt is the problem statement followed by the instruction ``Let's think step by step and output
the final answer within \textbackslash boxed\{\}.'', rendered with the model's chat template: the Llama-3 template used during SFT
for our 1.2B models, and the default Qwen3 template, with thinking enabled, for Qwen3. The ground-truth answer is the boxed answer
of the reference solution. Prompts longer than 1,024 tokens are discarded, which removes none of the 35,000 problems for our 1.2B
models and 2 for Qwen3.

\paragraph{Algorithm.}
We use GRPO~\citep{deepseek-math} as implemented in veRL~\citep{sheng2024hybridflow}. Each step samples 256 prompts and 4
rollouts per prompt at temperature 1.0 (no top-$k$ or top-$p$ truncation), with up to 3,072 generated tokens, using vLLM, with binary rewards. We add a KL penalty to the initial policy with coefficient $0.001$. %
We train for 2 epochs, or 272 steps total. %

\paragraph{Optimizers.}
All GRPO runs use a constant learning rate with no warmup or decay, and global gradient-norm clipping at $1.0$. AdamW uses
$(\beta_1, \beta_2) = (0.9, 0.999)$, $\epsilon = 10^{-8}$, and weight decay $0.01$. Muon follows Algorithm~\ref{alg:muon}, with
Nesterov momentum $\mu = 0.95$, five Newton--Schulz iterations, the update scaled by $0.2\sqrt{\max(m,n)}$, and weight decay
$0.01$. Embeddings, the language-model head, and
normalization parameters are optimized by an AdamW group with the same settings as the AdamW runs and the same learning rate as
the Muon matrices.

\paragraph{Learning-rate selection.}
We sweep the learning rate separately for each starting model and GRPO optimizer, in factor-of-two steps for our 1.2B models
and factor-$\sqrt{2}$ steps around the optimum for Qwen3. 
We select the learning rate with the highest
final-step accuracy on the 4,500 problems that are not in MATH-500~\citep{lightman2024lets}, so the selection is independent of
our MATH-500 results. Figures~\ref{fig:grpo_1p2b_lr_sweep} and~\ref{fig:grpo_qwen3_lr_sweep} show the results of the learning rate sweeps.
For each selected learning rate, we train three runs with different seeds.

\begin{table}[h]
\centering
\caption{Best GRPO learning rate (256 prompts $\times$ 4 rollouts), selected by final validation accuracy on the 4,500 MATH test
problems outside MATH-500.}
\label{tab:grpo_best_lr}
\begin{tabular}{lcc}
\toprule
Starting model & AdamW & Muon \\
\midrule
1.2B AdamW-PT, AdamW-SFT (8$\times$) & $2 \times 10^{-6}$ & $4 \times 10^{-6}$ \\
1.2B Muon-PT, Muon-SFT (8$\times$) & $4 \times 10^{-6}$ & $1.6 \times 10^{-5}$ \\
Qwen3-1.7B & $1.41 \times 10^{-6}$ & $5.66 \times 10^{-6}$ \\
Qwen3-4B & $1.41 \times 10^{-6}$ & $4 \times 10^{-6}$ \\
\bottomrule
\end{tabular}
\end{table}

\paragraph{Evaluation.}
We evaluate the final checkpoint (step 272) of each run on MATH-500 and on OlympiadBench~\citep{he-etal-2024-olympiadbench}, using its
674 open-ended, text-only, English competition-math problems (\texttt{OE\_TO\_maths\_en\_COMP}), with greedy decoding, up to 3,072 generated tokens,
the prompt format from training, and the training reward function for scoring. %

\paragraph{Forgetting.}
All forgetting measurements are differences between the final GRPO checkpoint and the model before GRPO.
\begin{itemize}
  \item \textbf{Pretraining validation loss} (1.2B models only): full-sequence cross-entropy on 26M tokens (800 batches of
  $8 \times 4096$-token windows) of held-out pretraining-mixture documents. These documents appear in neither the
  1$\times$ nor the 8$\times$ pretraining data.%
  \item \textbf{SFT validation loss} (1.2B models only): assistant-token cross-entropy on the 2,000 held-out Tulu-3 conversations
  used for SFT validation. %
  \item \textbf{6-task average}: following \citet{shenfeld2026rls}, we evaluate HellaSwag (normalized accuracy), TruthfulQA
  (MC2), MMLU (accuracy), IFEval (prompt-level strict accuracy), Winogrande (accuracy), and HumanEval (pass@1) with
  lm-evaluation-harness v0.4.9~\citep{eval-harness}, zero-shot. IFEval and HumanEval are run with each model's chat template (Qwen3 thinking disabled), and the other four tasks without one. Context length is 4,096 tokens for our
  1.2B models and 8,192 tokens for Qwen3.
\end{itemize}

\section{Additional Results: SFT}

\subsection{SFT benchmark performance for 1.2B models}

\begin{table*}[t]
\centering
\caption{\textbf{SFT benchmark performance at the best learning rate for each setting.} Average and standard deviation are reported over 3 random SFT seeds.}
\label{tab:sft_grid}
\small
\setlength{\tabcolsep}{4pt}
\begin{tabular}{@{}lccccccc@{}}
\toprule
\textbf{PT $\backslash$ SFT}
& \textbf{AdamW-SFT}
& \textbf{p0.1}
& \textbf{p0.25}
& \textbf{p0.5}
& \textbf{p0.75}
& \textbf{p0.9}
& \textbf{Muon-SFT} \\
\midrule

\textbf{AdamW-PT}
& $15.8 \pm 0.4$
& $15.8 \pm 0.5$
& $17.2 \pm 0.2$
& $17.0 \pm 0.4$
& $17.5 \pm 0.4$
& $17.4 \pm 0.4$
& $17.2 \pm 0.6$ \\

$p0.1$
& $16.8 \pm 0.8$
& $17.7 \pm 0.5$
& $18.0 \pm 0.2$
& $19.3 \pm 0.4$
& $18.3 \pm 0.5$
& $18.9 \pm 0.1$
& $18.9 \pm 0.3$ \\

$p0.25$
& $17.7 \pm 0.4$
& $17.2 \pm 0.2$
& $18.0 \pm 0.4$
& $17.9 \pm 0.2$
& $18.3 \pm 0.4$
& $18.0 \pm 0.3$
& $18.5 \pm 1.0$ \\

$p0.5$
& $17.0 \pm 0.4$
& $17.3 \pm 0.6$
& $17.5 \pm 0.4$
& $17.9 \pm 0.7$
& $18.3 \pm 0.4$
& $18.4 \pm 0.1$
& $19.0 \pm 0.4$ \\

$p0.75$
& $16.8 \pm 0.3$
& $17.0 \pm 0.7$
& $17.2 \pm 0.4$
& $17.9 \pm 0.1$
& $18.4 \pm 0.2$
& $18.9 \pm 0.2$
& $18.5 \pm 0.1$ \\

$p0.9$
& $17.3 \pm 0.7$
& $17.1 \pm 0.7$
& $17.7 \pm 0.1$
& $18.1 \pm 0.5$
& $17.8 \pm 0.9$
& $17.8 \pm 0.7$
& $18.6 \pm 0.5$ \\

\textbf{Muon-PT}
& $17.0 \pm 0.3$
& $17.0 \pm 0.5$
& $17.2 \pm 0.6$
& $18.2 \pm 0.6$
& $18.2 \pm 0.4$
& $18.6 \pm 0.3$
& $18.8 \pm 0.2$ \\

\bottomrule
\end{tabular}
\end{table*}

Table~\ref{tab:sft_grid} reports the average and standard deviation for benchmark averages across three random seeds for the SFT experiments in Section~\ref{sec:cost_of_optimizer_mismatch}. These are the results used to calculate row-wise averages, column-wise averages, and residuals in Figure~\ref{fig:residual_heatmap} (left).

\begin{figure}
    \centering
    \includegraphics[width=0.6\linewidth]{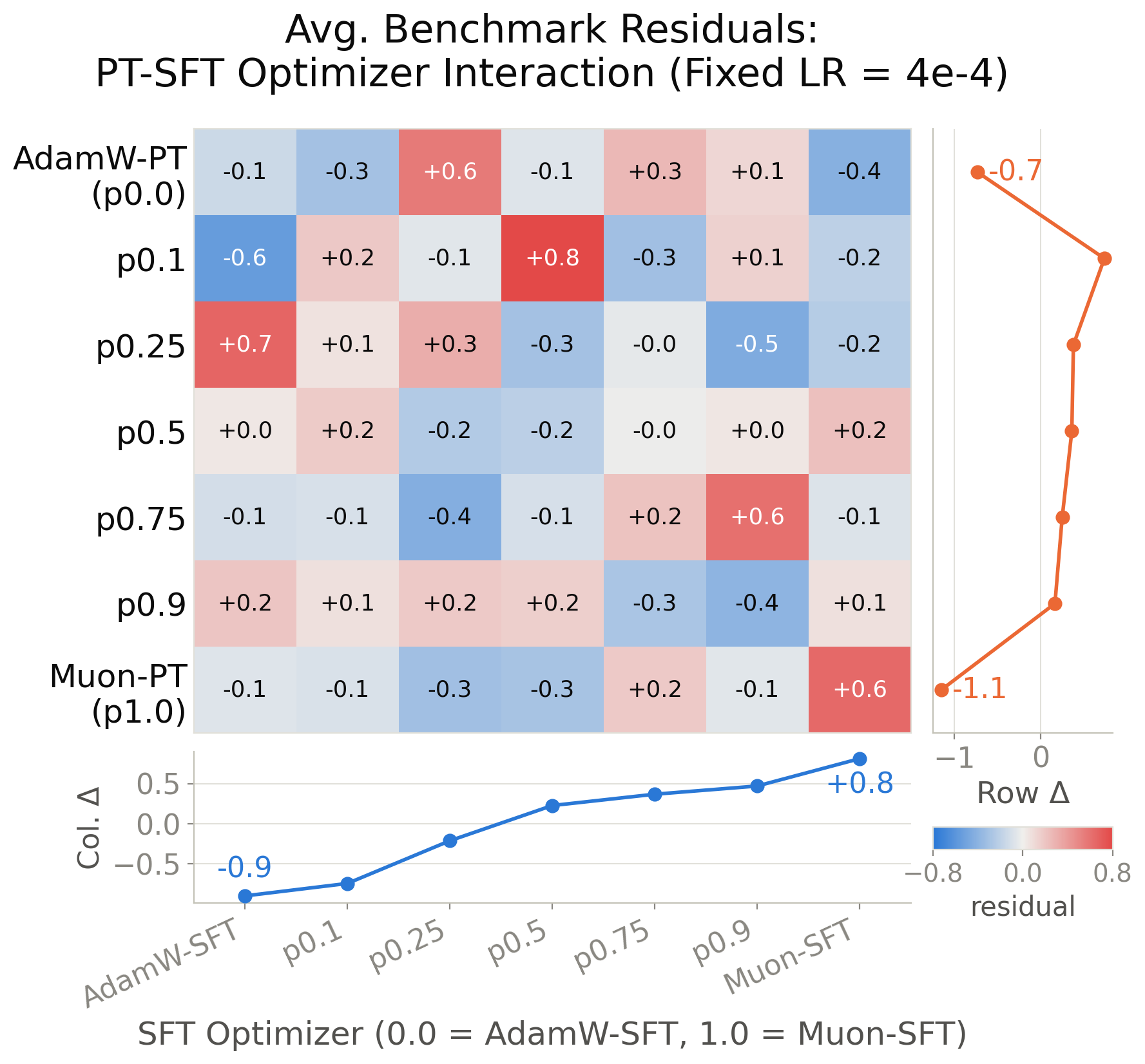}
    \caption{\textbf{SFT benchmark performance at uniform LR 4e-4.} Three independent SFT runs are averaged for each cell.}
    \label{fig:residual_heatmap_cpt_sft_uniform_lr}
\end{figure}

\subsection{SFT benchmark performance at uniform LR for 1.2B models}

As a companion to Figure~\ref{fig:residual_heatmap}, we recreate the same heatmap using models trained at the same LR (4e-4) for all settings, which is the majority-optimal setting, and present the results in Figure~\ref{fig:residual_heatmap_cpt_sft_uniform_lr}. These results largely reflect the trends seen in Figure~\ref{fig:residual_heatmap}: we see that increasing the proportion of Muon updates used in the SFT stage generally increases final benchmark performance. In the residual cells, we observe only mild interaction effects caused by optimizer consistency across training stages.

\begin{figure}
  \centering
  \begin{minipage}[t]{0.33\linewidth}
    \centering
    \includegraphics[width=\linewidth]{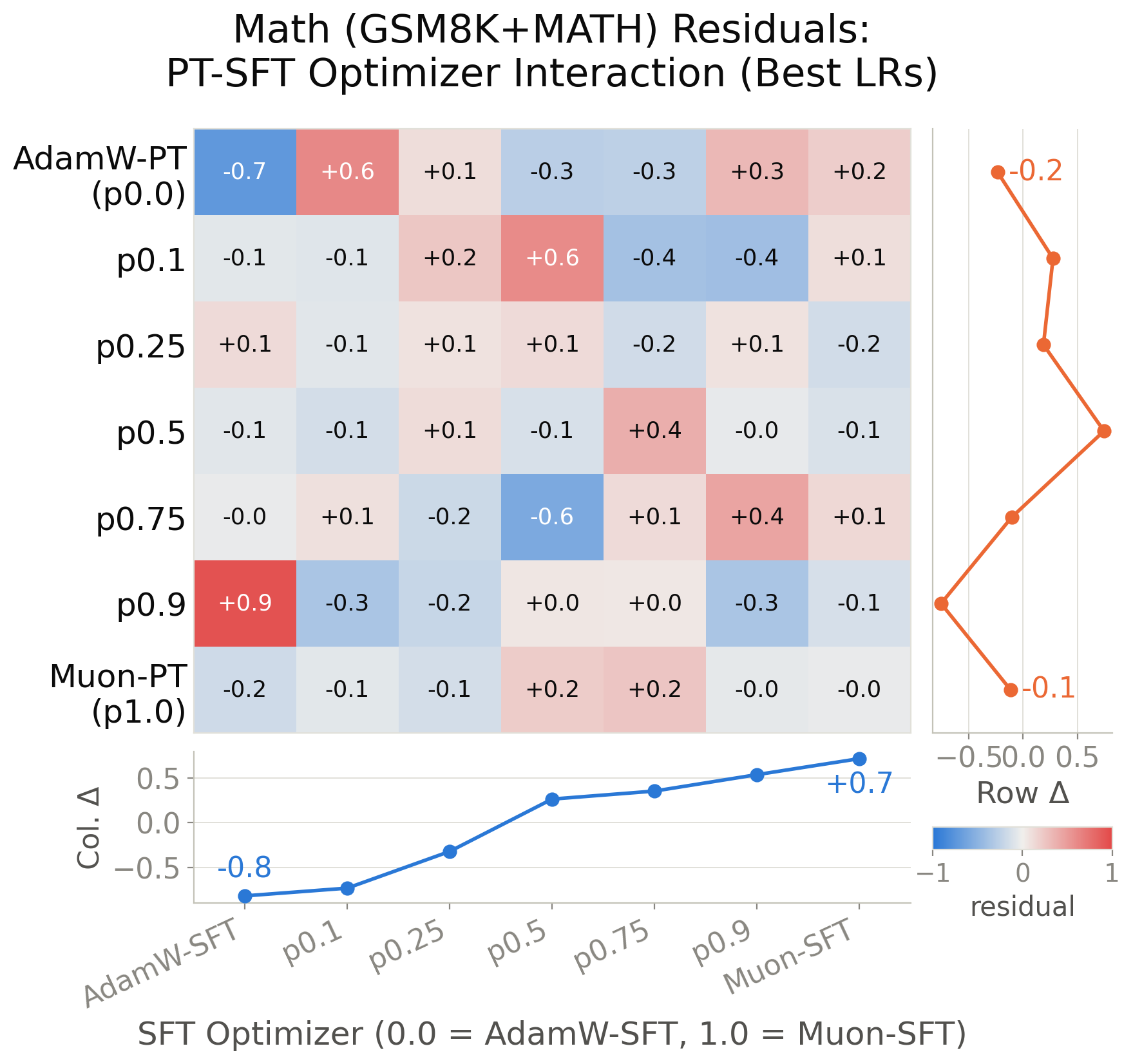}
  \end{minipage}\hfill
  \begin{minipage}[t]{0.33\linewidth}
    \centering
    \includegraphics[width=\linewidth]{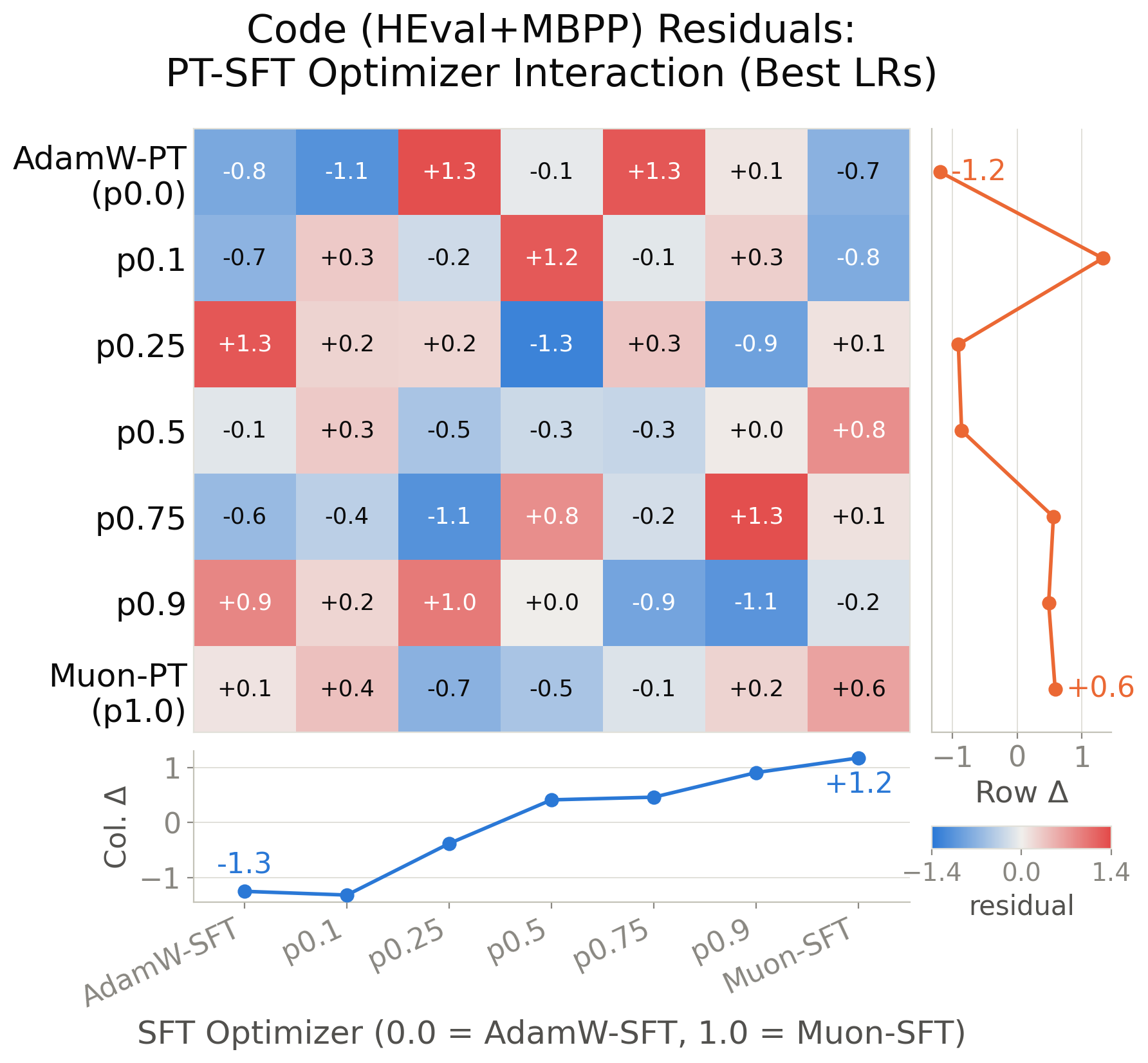}
  \end{minipage}\hfill
  \begin{minipage}[t]{0.33\linewidth}
    \centering
    \includegraphics[width=\linewidth]{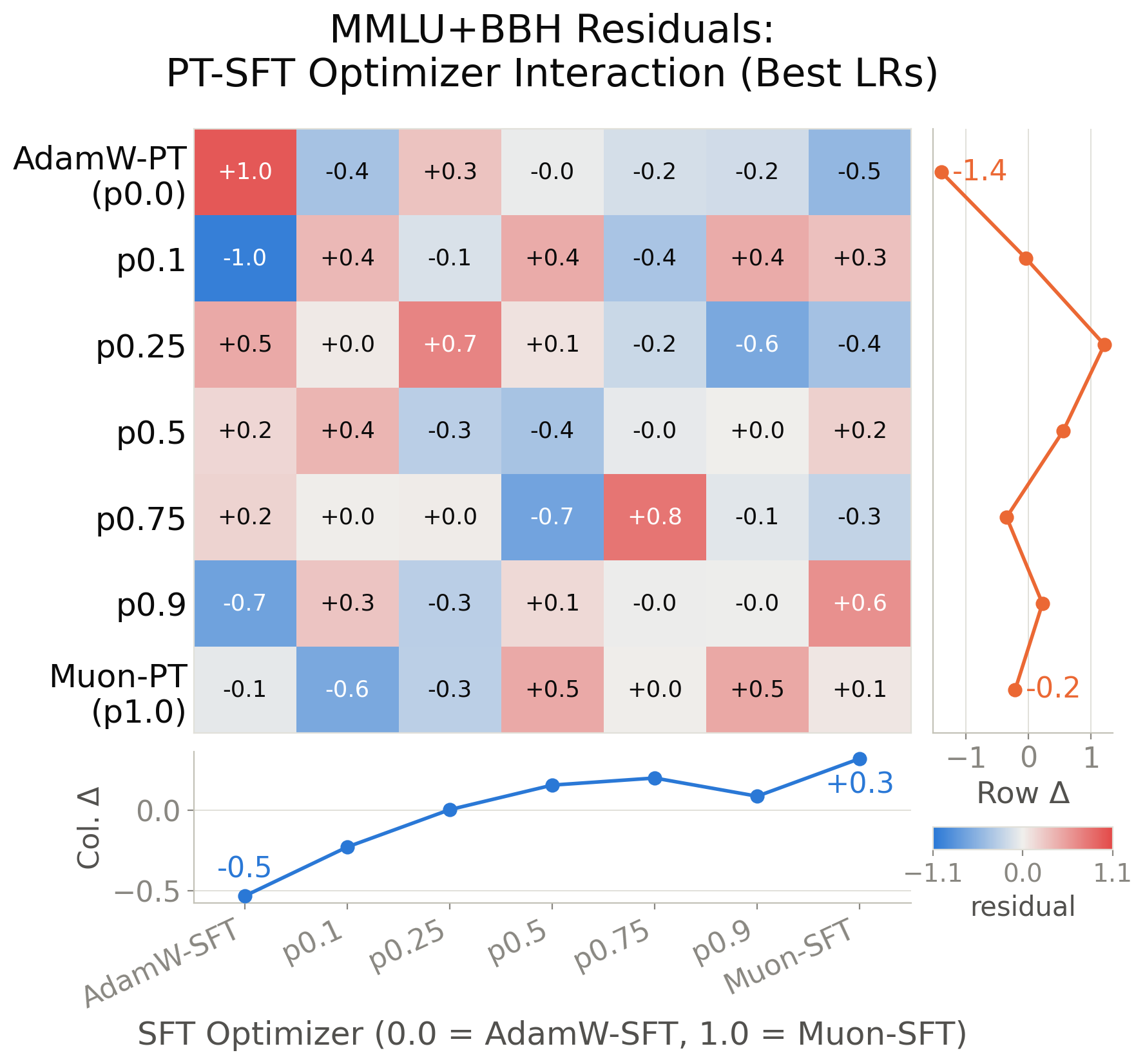}
  \end{minipage}
    \caption{\textbf{SFT benchmark performance by task type at the best LR for each setting.} Breakdown of the SFT results in Figure~1 by downstream task category for our 1.2B Llama-3 models. Rows index the proportion of Muon updates used during pretraining ($0.0=\mathrm{AdamW}$, $1.0=\mathrm{Muon}$), and columns index the proportion used during SFT. Results show the same qualitative trends as the aggregate benchmark performance in Figure~1: increasing the proportion of Muon updates during SFT generally improves performance across task types, with relatively small interaction effects between the pretraining and SFT optimizer choices.}
  \label{fig:residual_heatmap_by_task}
\end{figure}

\subsection{SFT benchmark performance by task type for 1.2B models}
\label{app:sft_by_task_type}

Figure~\ref{fig:residual_heatmap_by_task} breaks down the SFT results from Figure~\ref{fig:residual_heatmap} by benchmark type rather than averaging across all six downstream tasks. Across task categories, we observe the same qualitative trends as in the aggregate results: performance is largely explained by the independent effects of the pretraining and SFT optimizer choices, with relatively small interaction effects from matching the optimizer across stages. In particular, increasing the proportion of Muon updates during SFT generally improves performance across task types, regardless of the optimizer distribution used during pretraining.

\subsection{Post-SFT forgetting for 1.2B models}

\begin{table*}[t]
\centering
\caption{\textbf{Forgetting after SFT @ LR 4e-4, as measured by change in pretraining validation loss.} Average and standard deviation are reported over 3 random seeds. These values are used to calculate the residuals in Figure~\ref{fig:residual_heatmap} (right). A smaller value indicates less forgetting and indicates that the model has retained more pretraining knowledge.}
\label{tab:forget_grid}
\scriptsize
\setlength{\tabcolsep}{4pt}
\begin{tabular}{@{}lccccccc@{}}
\toprule
\textbf{PT $\backslash$ SFT}
& \textbf{AdamW}
& \textbf{0.1}
& \textbf{0.25}
& \textbf{0.5}
& \textbf{0.75}
& \textbf{0.9}
& \textbf{Muon} \\
\midrule

\textbf{AdamW}
& $+0.49 \pm 0.01$
& $+0.54 \pm 0.02$
& $+0.56 \pm 0.02$
& $+0.84 \pm 0.37$
& $+0.83 \pm 0.18$
& $+0.75 \pm 0.09$
& $+0.73 \pm 0.02$ \\

$p0.1$
& $+0.52 \pm 0.10$
& $+0.58 \pm 0.05$
& $+0.64 \pm 0.09$
& $+0.63 \pm 0.06$
& $+0.71 \pm 0.09$
& $+0.80 \pm 0.07$
& $+0.67 \pm 0.04$ \\

$p0.25$
& $+0.42 \pm 0.02$
& $+0.45 \pm 0.03$
& $+0.49 \pm 0.01$
& $+0.51 \pm 0.01$
& $+0.59 \pm 0.01$
& $+0.61 \pm 0.01$
& $+0.60 \pm 0.01$ \\

$p0.5$
& $+0.40 \pm 0.02$
& $+0.42 \pm 0.04$
& $+0.44 \pm 0.02$
& $+0.49 \pm 0.01$
& $+0.54 \pm 0.02$
& $+0.57 \pm 0.04$
& $+0.58 \pm 0.02$ \\

$p0.75$
& $+0.38 \pm 0.02$
& $+0.44 \pm 0.03$
& $+0.46 \pm 0.03$
& $+0.46 \pm 0.03$
& $+0.51 \pm 0.03$
& $+0.53 \pm 0.02$
& $+0.54 \pm 0.00$ \\

$p0.9$
& $+0.46 \pm 0.05$
& $+0.65 \pm 0.27$
& $+0.77 \pm 0.21$
& $+0.54 \pm 0.08$
& $+0.58 \pm 0.09$
& $+0.57 \pm 0.06$
& $+0.53 \pm 0.04$ \\

\textbf{Muon}
& $+0.36 \pm 0.01$
& $+0.37 \pm 0.02$
& $+0.39 \pm 0.00$
& $+0.36 \pm 0.00$
& $+0.39 \pm 0.04$
& $+0.39 \pm 0.03$
& $+0.38 \pm 0.00$ \\

\bottomrule
\end{tabular}
\end{table*}

Table~\ref{tab:forget_grid} reports the average and standard deviation for change in pretraining validation loss across three random seeds after SFT on our 1.2B Llama-3 models. These are the results used to calculate marginal and residual effects in Figure~\ref{fig:residual_heatmap} (right).

\section{Additional Results: CPT}

\begin{table*}[t]
\centering
\caption{\textbf{Continued pretraining (CPT) performance at LR 5e-5.} Average and standard deviation are reported over 3 random seeds. These values are used to calculate the residuals in Figure~\ref{fig:residual_heatmap_cpt} (left).}
\label{tab:cpt_grid_unif}
\scriptsize
\setlength{\tabcolsep}{2pt}
\begin{tabular}{@{}lccccccc@{}}
\toprule
\textbf{PT $\backslash$ CPT}
& \textbf{AdamW}
& \textbf{p0.1}
& \textbf{p0.25}
& \textbf{p0.5}
& \textbf{p0.75}
& \textbf{p0.9}
& \textbf{Muon} \\
\midrule

\textbf{AdamW}
& $2.6592 \pm .0001$
& $2.6587 \pm .0001$
& $2.6585 \pm .0001$
& $2.6583 \pm .0000$
& $2.6582 \pm .0001$
& $2.6581 \pm .0000$
& $2.6580 \pm .0000$ \\

$p0.1$
& $2.6225 \pm .0001$
& $2.6221 \pm .0001$
& $2.6219 \pm .0000$
& $2.6217 \pm .0001$
& $2.6215 \pm .0001$
& $2.6215 \pm .0000$
& $2.6214 \pm .0000$ \\

$p0.25$
& $2.6227 \pm .0000$
& $2.6223 \pm .0000$
& $2.6221 \pm .0000$
& $2.6218 \pm .0000$
& $2.6217 \pm .0000$
& $2.6216 \pm .0000$
& $2.6215 \pm .0000$ \\

$p0.5$
& $2.6190 \pm .0000$
& $2.6186 \pm .0000$
& $2.6184 \pm .0000$
& $2.6181 \pm .0000$
& $2.6179 \pm .0000$
& $2.6178 \pm .0001$
& $2.6177 \pm .0000$ \\

$p0.75$
& $2.6195 \pm .0000$
& $2.6191 \pm .0001$
& $2.6188 \pm .0000$
& $2.6185 \pm .0001$
& $2.6182 \pm .0000$
& $2.6181 \pm .0000$
& $2.6179 \pm .0000$ \\

$p0.9$
& $2.6175 \pm .0001$
& $2.6171 \pm .0000$
& $2.6168 \pm .0000$
& $2.6165 \pm .0001$
& $2.6162 \pm .0001$
& $2.6160 \pm .0000$
& $2.6159 \pm .0001$ \\

\textbf{Muon-PT}
& $2.6116 \pm .0000$
& $2.6112 \pm .0001$
& $2.6110 \pm .0001$
& $2.6108 \pm .0000$
& $2.6106 \pm .0000$
& $2.6104 \pm .0000$
& $2.6101 \pm .0000$ \\

\bottomrule
\end{tabular}
\end{table*}

\begin{table*}[t]
\centering
\caption{\textbf{Continued pretraining (CPT) performance at best LR for each setting.} Average and standard deviation are reported over 3 random seeds. These values are used to calculate the residuals in Figure~\ref{fig:residual_heatmap_cpt} (right).}
\label{tab:cpt_grid_best}
\scriptsize
\setlength{\tabcolsep}{2pt}
\begin{tabular}{@{}lccccccc@{}}
\toprule
\textbf{PT $\backslash$ CPT }
& \textbf{AdamW}
& \textbf{p0.1}
& \textbf{p0.25}
& \textbf{p0.5}
& \textbf{p0.75}
& \textbf{p0.9}
& \textbf{Muon} \\
\midrule

\textbf{AdamW}
& $2.6592 \pm .0001$
& $2.6587 \pm .0001$
& $2.6585 \pm .0001$
& $2.6583 \pm .0000$
& $2.6581 \pm .0000$
& $2.6581 \pm .0000$
& $2.6580 \pm .0000$ \\

$p0.1$
& $2.6225 \pm .0001$
& $2.6221 \pm .0001$
& $2.6219 \pm .0000$
& $2.6217 \pm .0001$
& $2.6215 \pm .0001$
& $2.6215 \pm .0000$
& $2.6214 \pm .0000$ \\

$p0.25$
& $2.6225 \pm .0000$
& $2.6222 \pm .0001$
& $2.6221 \pm .0000$
& $2.6218 \pm .0000$
& $2.6217 \pm .0000$
& $2.6216 \pm .0000$
& $2.6215 \pm .0000$ \\

$p0.5$
& $2.6187 \pm .0000$
& $2.6184 \pm .0000$
& $2.6183 \pm .0000$
& $2.6181 \pm .0000$
& $2.6179 \pm .0000$
& $2.6178 \pm .0001$
& $2.6177 \pm .0000$ \\

$p0.75$
& $2.6192 \pm .0003$
& $2.6186 \pm .0001$
& $2.6185 \pm .0000$
& $2.6183 \pm .0001$
& $2.6182 \pm .0000$
& $2.6181 \pm .0000$
& $2.6179 \pm .0000$ \\

$p0.9$
& $2.6171 \pm .0000$
& $2.6166 \pm .0000$
& $2.6164 \pm .0000$
& $2.6162 \pm .0001$
& $2.6160 \pm .0001$
& $2.6160 \pm .0000$
& $2.6158 \pm .0000$ \\

\textbf{Muon}
& $2.6114 \pm .0000$
& $2.6110 \pm .0000$
& $2.6108 \pm .0000$
& $2.6106 \pm .0000$
& $2.6106 \pm .0000$
& $2.6104 \pm .0000$
& $2.6099 \pm .0000$ \\

\bottomrule
\end{tabular}
\end{table*}

Table~\ref{tab:cpt_grid_unif} (uniform LR) and Table~\ref{tab:cpt_grid_best} (best LRs)
 report the average and standard deviation for CPT final loss across three random seeds for the CPT experiments in Section~\ref{sec:cost_of_optimizer_mismatch}. These are the results used to calculate row-wise averages, column-wise averages, and residuals in Figure~\ref{fig:residual_heatmap_cpt}.

\section{Additional Results: RLVR}
\label{app:rlvr}

\begin{table}[h!]
    \centering
    \caption{AdamW update RMS at different steps for 1.2B 8$\times$ Chinchilla and Qwen3 models.}
    \label{tab:grpo_adam_rms}
\small
\setlength{\tabcolsep}{8pt}
\begin{tabular}{@{}lcccc@{}}
\toprule
\textbf{Step}
& \textbf{8X AdamW-PT}
& \textbf{8X Muon-PT}
& \textbf{Qwen3-1.7B}
& \textbf{Qwen3-4B} \\
\midrule

25
& 0.248
& 0.246
& 0.240
& 0.225 \\

50
& 0.270
& 0.248
& 0.218
& 0.209 \\

100
& 0.226
& 0.238
& 0.205
& 0.227 \\

150
& 0.228
& 0.224
& 0.209
& 0.199 \\

200
& 0.219
& 0.215
& 0.197
& 0.186 \\

272
& 0.235
& 0.224
& 0.227
& 0.200 \\

\bottomrule
\end{tabular}
\end{table}

\begin{table}[h!]
    \centering
    \caption{Muon update RMS at different steps for 1.2B 8$\times$ Chinchilla and Qwen3 models.}
    \label{tab:grpo_muon_rms}
\small
\setlength{\tabcolsep}{8pt}
\begin{tabular}{@{}lcccc@{}}
\toprule
\textbf{Step}
& \textbf{8X AdamW-PT}
& \textbf{8X Muon-PT}
& \textbf{Qwen3-1.7B}
& \textbf{Qwen3-4B} \\
\midrule

25
& 0.1664
& 0.1784
& 0.1826
& 0.1827 \\

50
& 0.1665
& 0.1789
& 0.1829
& 0.1825 \\

100
& 0.1667
& 0.1794
& 0.1828
& 0.1831 \\

150
& 0.1668
& 0.1790
& 0.1826
& 0.1828 \\

200
& 0.1668
& 0.1789
& 0.1828
& 0.1831 \\

272
& 0.1670
& 0.1787
& 0.1827
& 0.1819 \\

\bottomrule
\end{tabular}
\end{table}

As shown in Appendix~\ref{app:rlvr_hyperparameters}, Muon consistently prefers a higher learning rate than AdamW for the same base model. This differs from the pattern in the SFT setting (Appendix~\ref{app:sft_hyperparameters}), where we find that Muon prefers an equal or \textbf{lower} learning rate than AdamW. To investigate this, we first measure the empirical update size of both optimizers, and then compare the sparsity, Frobenius norm, and operator norms of the resulting weight changes from GRPO.

\paragraph{Muon prefers larger learning rates in part due to smaller update RMS.}
Since our GRPO experiments use $\beta_1 = 0.9$ for AdamW, the expected update RMS is approximately $0.2$ \citep{su2025adamrms}. Muon is similarly scaled to achieve an update RMS of $0.2$ under ideal orthogonalization. However, empirical update magnitudes may differ due to inexact orthogonalization and deviations from theoretical assumptions. We therefore measure update RMS throughout training in Tables~\ref{tab:grpo_adam_rms} and~\ref{tab:grpo_muon_rms}. AdamW's update RMS is approximately $1.3$--$1.4\times$ larger for our 1.2B models, and $1.1$--$1.2\times$ larger for Qwen3. These differences are substantially smaller than the $2$--$4\times$ ratio of optimal learning rates, suggesting that update normalization alone cannot explain Muon's preference for larger learning rates during GRPO.

\paragraph{Muon and AdamW produce different weight change geometries.}
Prior work has examined how optimizer choice affects update sparsity \citep{mukherjee2026needadamsurprisinglystrong} and matrix norms \citep{liu2026optimizer}, potentially influencing forgetting. We examine the accumulated parameter change $\Delta W = W_T - W_0$, where $W_0$ and $W_T$ denote the model parameters before and after $T$ GRPO steps.

\begin{figure}[t]
    \centering
    \includegraphics[width=1\linewidth, trim=0 13.5cm 0 0, clip]{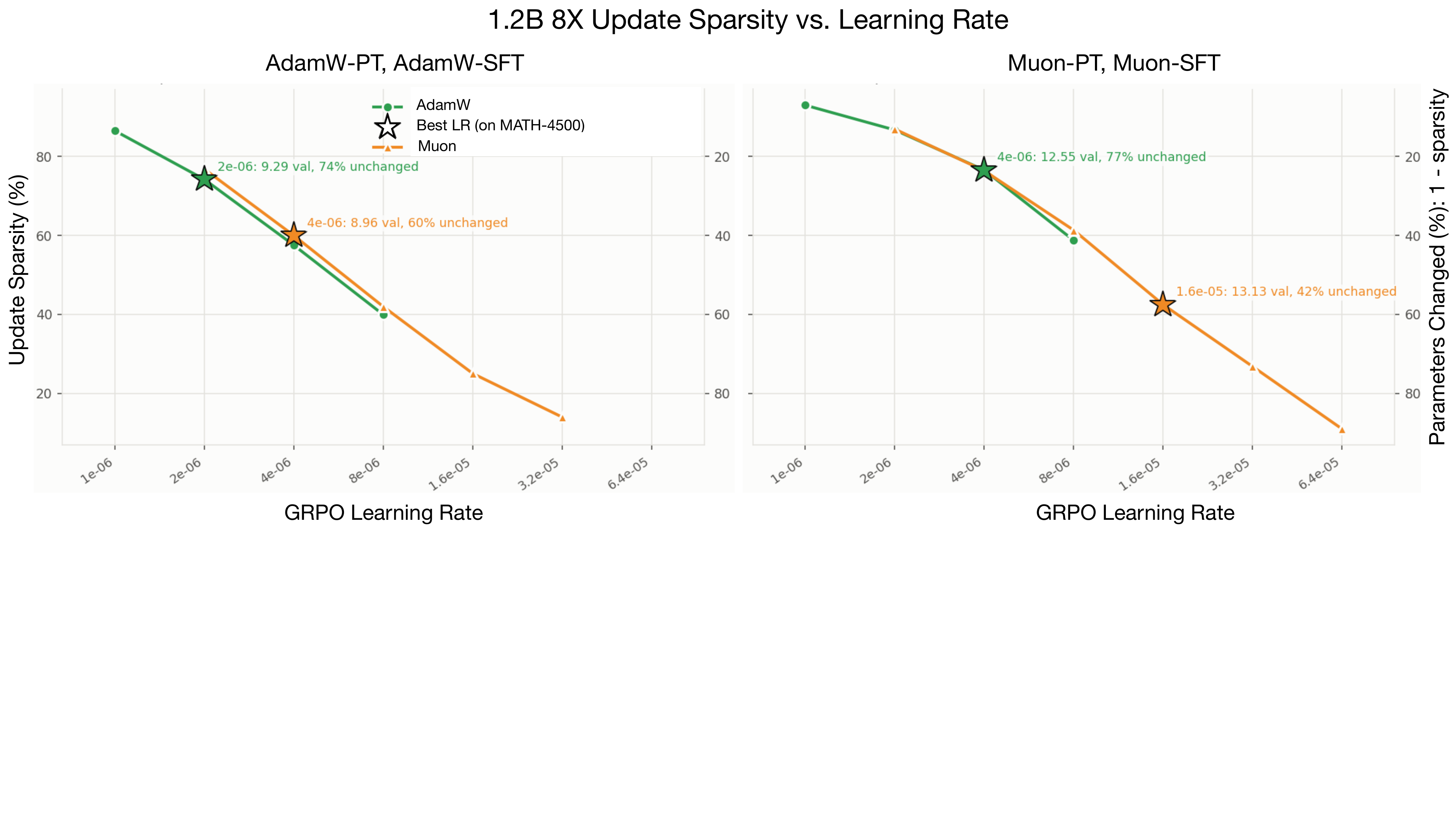}
    \caption{
        \textbf{Parameter sparsity after GRPO for 1.2B models.}
        Fraction of approximately unchanged parameters
        ($|\Delta W_i| \leq 10^{-5}$) after 272 GRPO steps,
        as a function of learning rate, for our 1.2B
        8$\times$ Chinchilla models. Stars indicate
        the selected learning rates. Lower sparsity
        indicates denser parameter changes.
    }
    \label{fig:grpo_1p2b_sparsity}
\end{figure}

\begin{figure}[t!]
    \centering
    \includegraphics[width=1.0\linewidth, trim=0 13.5cm 0 0, clip]{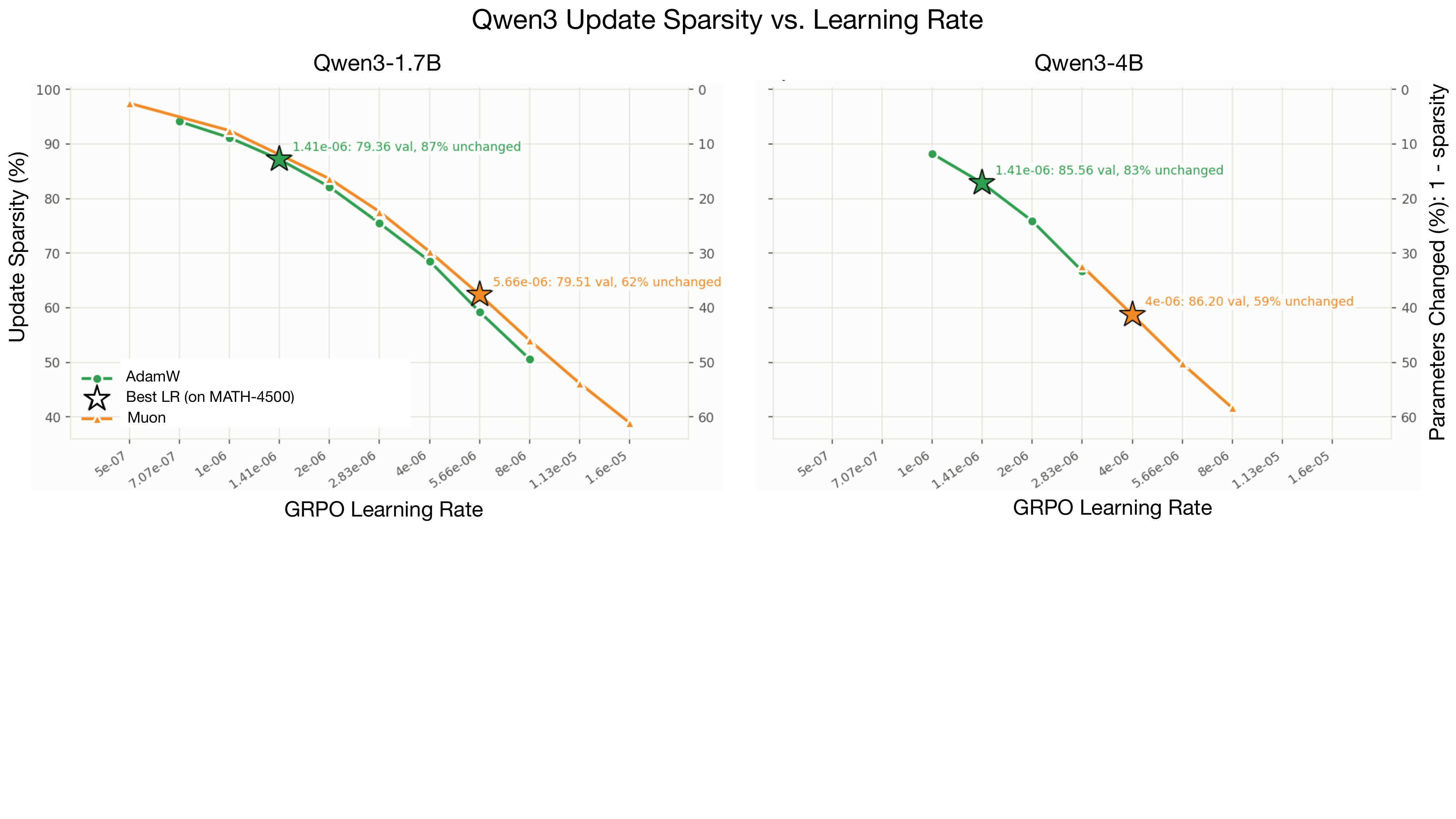}
    \caption{
        \textbf{Parameter sparsity after GRPO for Qwen3 models.}
        Fraction of approximately unchanged parameters
        after GRPO for Qwen3-1.7B and 4B, following
        Figure~\ref{fig:grpo_1p2b_sparsity}.
        Muon produces denser parameter changes at
        its selected learning rate for both models.
    }
    \label{fig:grpo_qwen3_sparsity}
\end{figure}

In Figures~\ref{fig:grpo_1p2b_sparsity} and~\ref{fig:grpo_qwen3_sparsity}, we measure the fraction of approximately unchanged parameters,
\begin{equation}
    S_\epsilon(\Delta W)
    = \frac{1}{N}\sum_{i=1}^{N}
    \mathbf{1}\left[|\Delta W_i| \leq \epsilon\right],
    \qquad \epsilon = 10^{-5},
\end{equation}
where $N$ is the number of parameters. At their respective optimal learning rates, Muon consistently produces denser parameter changes than AdamW.

In Tables~\ref{tab:grpo_1p2b_parameter_changes} and~\ref{tab:grpo_qwen3_parameter_changes}, we further compare the Frobenius norm, operator norm, and stable rank $\|\Delta W\|_F^2/\|\Delta W\|_{\mathrm{op}}^2$, averaged across weight matrices. Despite producing larger Frobenius norms, Muon yields smaller operator norms and substantially higher stable ranks ($359$--$656$, versus $5$--$19$ for AdamW). This suggests that Muon's accumulated parameter changes are distributed across more singular directions, whereas AdamW's changes are concentrated in fewer dominant directions. Nevertheless, these substantially different weight change geometries do not translate into systematically greater forgetting (Table~\ref{tab:grpo_1b} and Table~\ref{tab:grpo_qwen3}).

\begin{table}[t]
    \centering
    \caption{\textbf{Geometry of accumulated parameter changes after GRPO for 1.2B models.}
    We compute $\Delta W = W_T - W_0$ and report
    matrix-wise averages over 224 attention and MLP
    weight matrices. Values are mean $\pm$ standard
    deviation over three seeds. Stable rank is
    $\|\Delta W\|_F^2/\|\Delta W\|_{\mathrm{op}}^2$.}
    \label{tab:grpo_1p2b_parameter_changes}
    \small
    \setlength{\tabcolsep}{6pt}
    \resizebox{\textwidth}{!}{%
    \begin{tabular}{@{}lccccc@{}}
    \toprule
    \textbf{Pretraining $\rightarrow$ GRPO (LR)}
    & $\|\Delta W\|_F$
    & $\|\Delta W\|_F / \|W_0\|_F$
    & $\|\Delta W\|_{\mathrm{op}}$
    & $\|\Delta W\|_{\mathrm{op}} / \|W_0\|_{\mathrm{op}}$
    & \textbf{Stable rank} \\
    \midrule

    AdamW $\rightarrow$ AdamW ($2\times10^{-6}$)
    & $0.0698 \pm 0.0011$
    & $(1.40 \pm 0.02)\times10^{-3}$
    & $0.0195 \pm 0.0001$
    & $(4.55 \pm 0.03)\times10^{-3}$
    & $13.0 \pm 0.3$ \\

    AdamW $\rightarrow$ Muon-RMS ($4\times10^{-6}$)
    & $0.1164 \pm 0.0002$
    & $(2.26 \pm 0.003)\times10^{-3}$
    & $0.0060 \pm 0.0001$
    & $(1.28 \pm 0.02)\times10^{-3}$
    & $359.1 \pm 10.2$ \\

    Muon-RMS $\rightarrow$ AdamW ($4\times10^{-6}$)
    & $0.1508 \pm 0.0032$
    & $(1.12 \pm 0.02)\times10^{-3}$
    & $0.0657 \pm 0.0021$
    & $(6.95 \pm 0.23)\times10^{-3}$
    & $5.4 \pm 0.1$ \\

    Muon-RMS $\rightarrow$ Muon-RMS ($1.6\times10^{-5}$)
    & $0.5111 \pm 0.0014$
    & $(3.67 \pm 0.01)\times10^{-3}$
    & $0.0235 \pm 0.00004$
    & $(2.40 \pm 0.004)\times10^{-3}$
    & $450.0 \pm 2.4$ \\

    \bottomrule
    \end{tabular}%
    }
\end{table}

\begin{table}[t]
    \centering
    \caption{\textbf{Geometry of accumulated parameter changes after GRPO for Qwen3 models.}
    }
    \label{tab:grpo_qwen3_parameter_changes}
    \small
    \setlength{\tabcolsep}{6pt}
    \resizebox{\textwidth}{!}{%
    \begin{tabular}{@{}lccccc@{}}
    \toprule
    \textbf{Model $\rightarrow$ GRPO (LR)}
    & $\|\Delta W\|_F$
    & $\frac{\|\Delta W\|_F}{\|W_0\|_F}$
    & $\|\Delta W\|_{\mathrm{op}}$
    & $\frac{\|\Delta W\|_{\mathrm{op}}}{\|W_0\|_{\mathrm{op}}}$
    & \textbf{Stable Rank} \\
    \midrule

    Qwen3-1.7B $\rightarrow$ AdamW ($1.41\times10^{-6}$)
    & $0.0583 \pm 0.0009$
    & $(6.87 \pm 0.11)\times10^{-4}$
    & $0.0149 \pm 0.0002$
    & $(2.15 \pm 0.02)\times10^{-3}$
    & $16.8 \pm 0.4$ \\

    Qwen3-1.7B $\rightarrow$ Muon-RMS ($5.66\times10^{-6}$)
    & $0.2128 \pm 0.0003$
    & $(2.46 \pm 0.003)\times10^{-3}$
    & $0.0092 \pm 0.00004$
    & $(1.32 \pm 0.004)\times10^{-3}$
    & $514.5 \pm 4.8$ \\

    \midrule

    Qwen3-4B $\rightarrow$ AdamW ($1.41\times10^{-6}$)
    & $0.0735 \pm 0.0004$
    & $(9.06 \pm 0.06)\times10^{-4}$
    & $0.0174 \pm 0.0005$
    & $(2.88 \pm 0.08)\times10^{-3}$
    & $19.0 \pm 0.8$ \\

    Qwen3-4B $\rightarrow$ Muon-RMS ($4\times10^{-6}$)
    & $0.2178 \pm 0.0030$
    & $(2.65 \pm 0.04)\times10^{-3}$
    & $0.0083 \pm 0.00005$
    & $(1.40 \pm 0.009)\times10^{-3}$
    & $656.2 \pm 20.2$ \\

    \bottomrule
    \end{tabular}%
    }
\end{table}

\begin{figure}[t]
    \centering
    \includegraphics[width=1.0\linewidth, trim=0 16cm 0 0, clip]{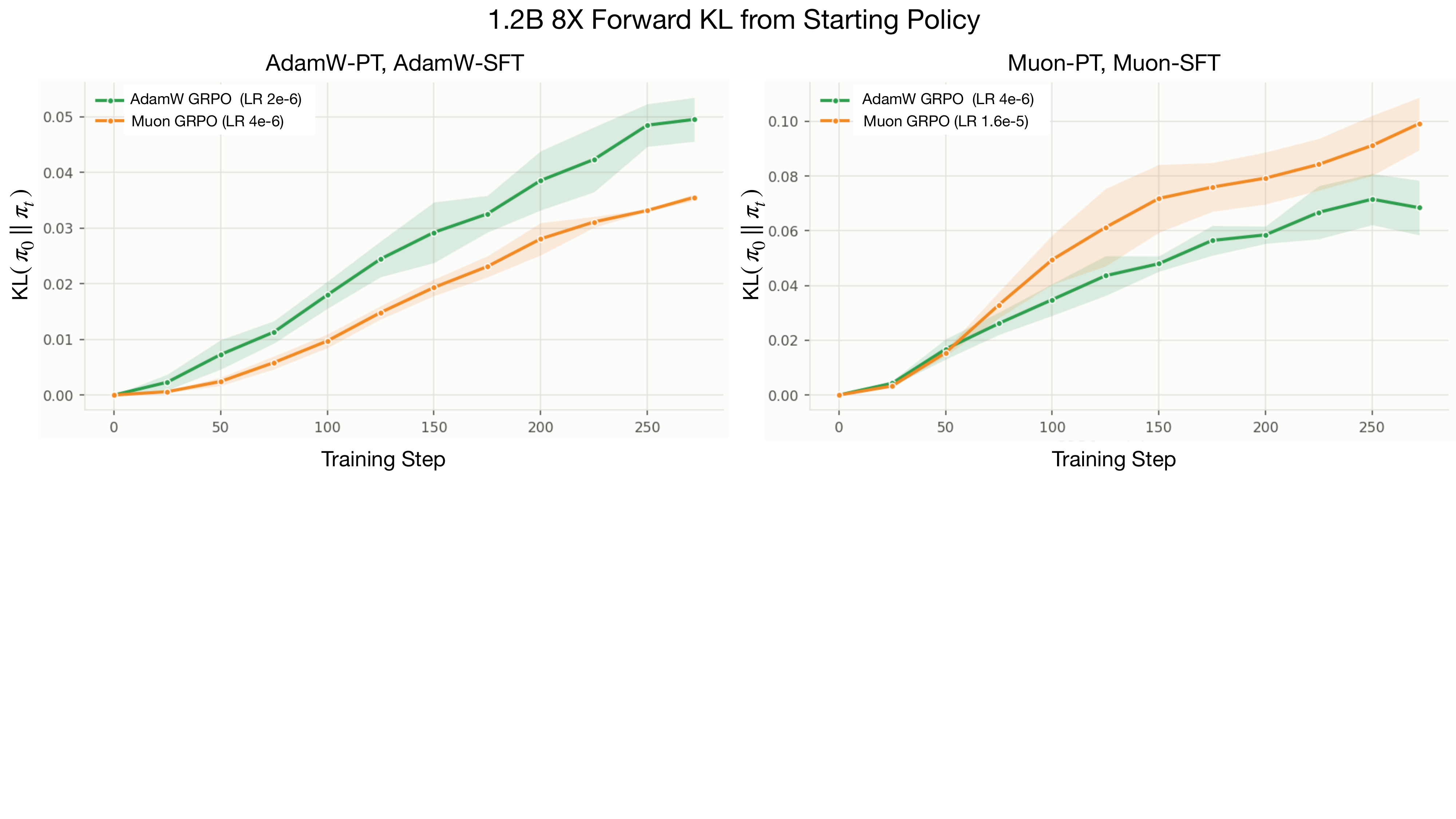}
    \caption{\textbf{Forward KL divergence during GRPO for 1.2B models.}
Forward KL from the initial policy, evaluated
on 1,000 held-out NuminaMath prompts with four
sampled responses each. The curves show the mean
over three seeds, with shaded regions indicating
$\pm 1$ standard deviation. Both optimizers
use their selected learning rates based on MATH-4500 performance.
}
    \label{fig:grpo_1p2b_traj_kl}
\end{figure}

\begin{figure}[t]
    \centering
    \includegraphics[width=1\linewidth, trim=0 16cm 0 0, clip]{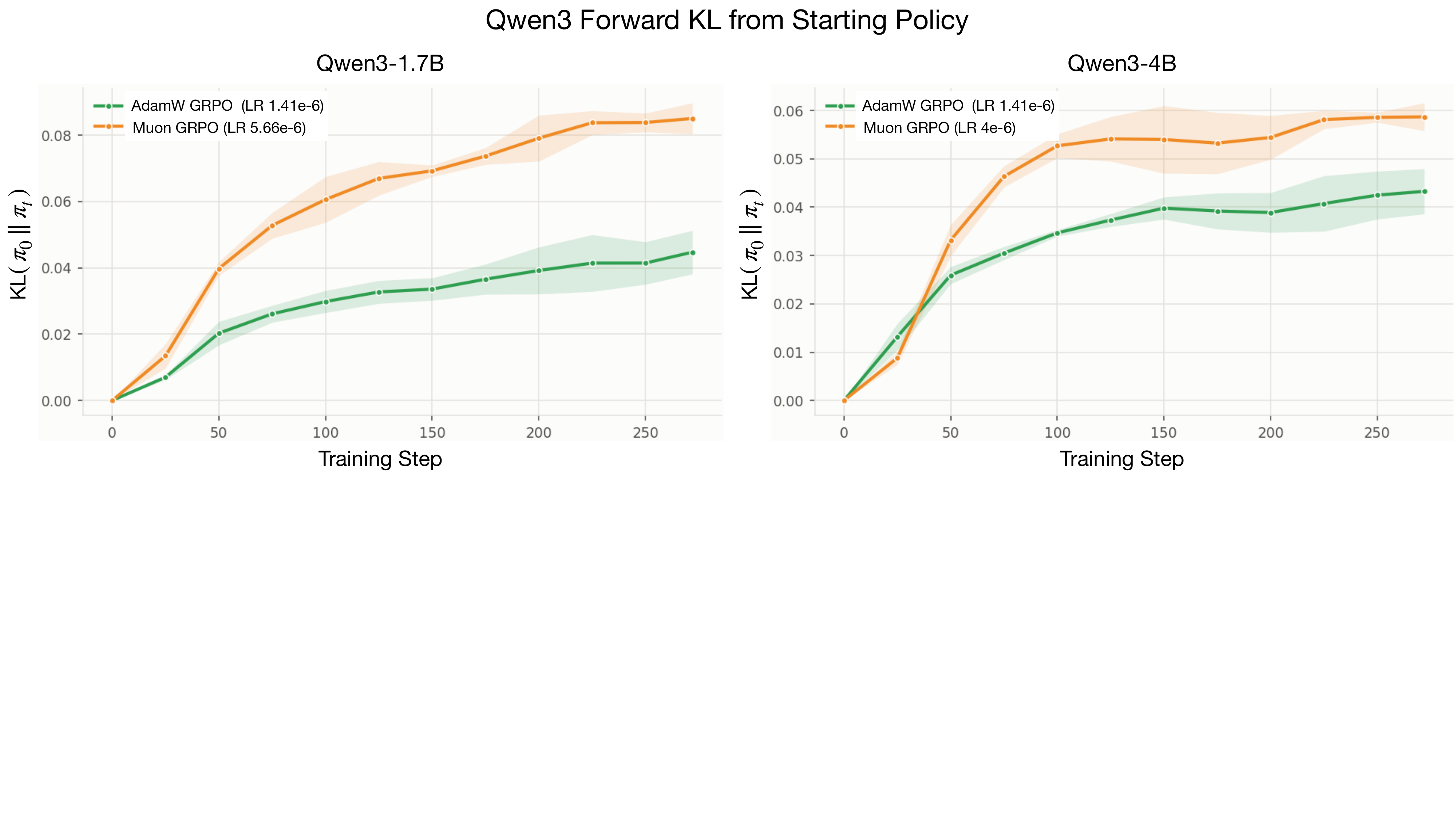}
    \caption{\textbf{Forward KL divergence during GRPO for Qwen3 models.}
Forward KL from the initial policy for
Qwen3-1.7B and 4B. Evaluation and formatting
follow Figure ~\ref{fig:grpo_1p2b_traj_kl}. Muon exhibits greater policy
drift than AdamW for both models.
}
    \label{fig:grpo_qwen3_traj_kl}
\end{figure}

\begin{figure}
    \centering
    \includegraphics[width=1\linewidth]{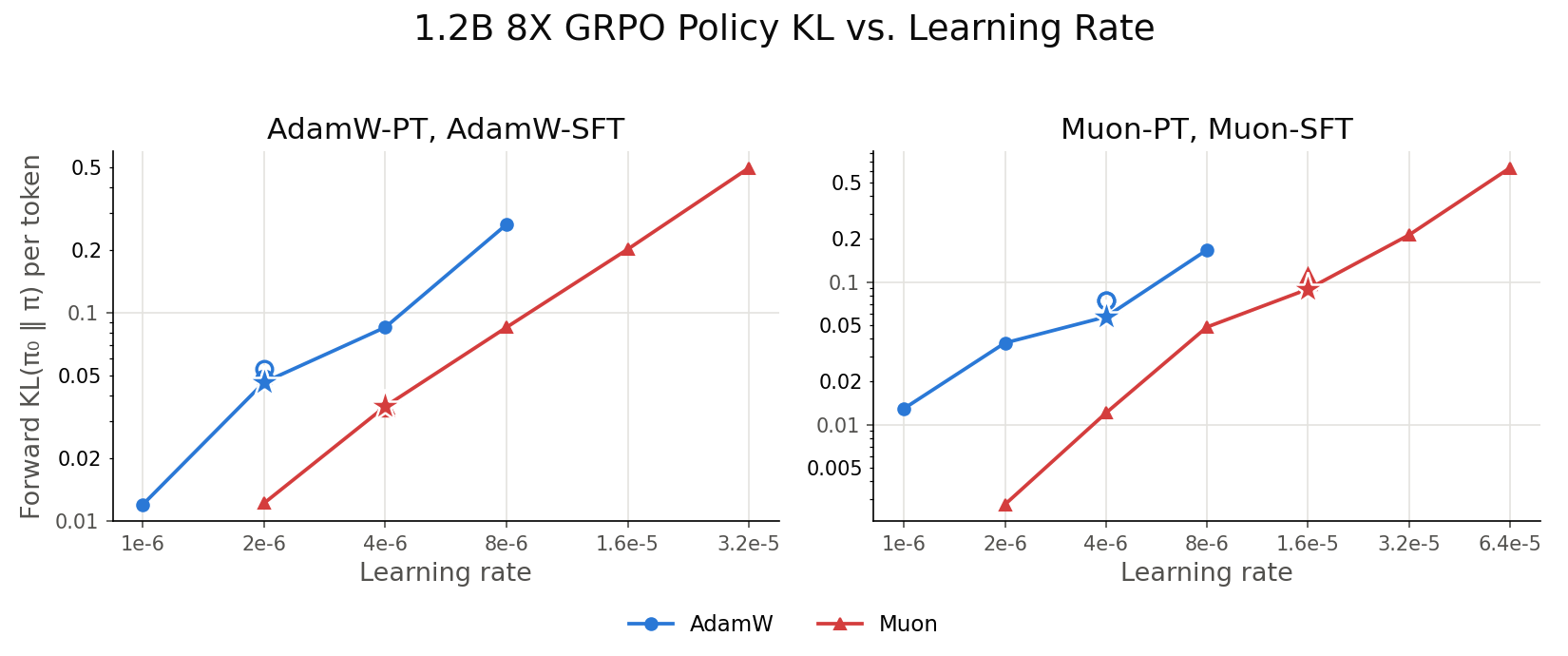}
    \caption{\textbf{Forward KL versus learning rate for 1.2B models.}
Final forward KL from the initial policy across GRPO learning-rate sweeps for our 1.2B 8$\times$ Chinchilla models. Stars indicate the selected learning rates based on MATH-4500 performance, and hollow markers indicate additional random seeds at those learning rates. AdamW produces greater policy drift than Muon at the same learning rate.}

    \label{fig:grpo_1p2b_kl_vs_lr}
\end{figure}

\begin{figure}[t]
    \centering
    \includegraphics[width=1\linewidth]{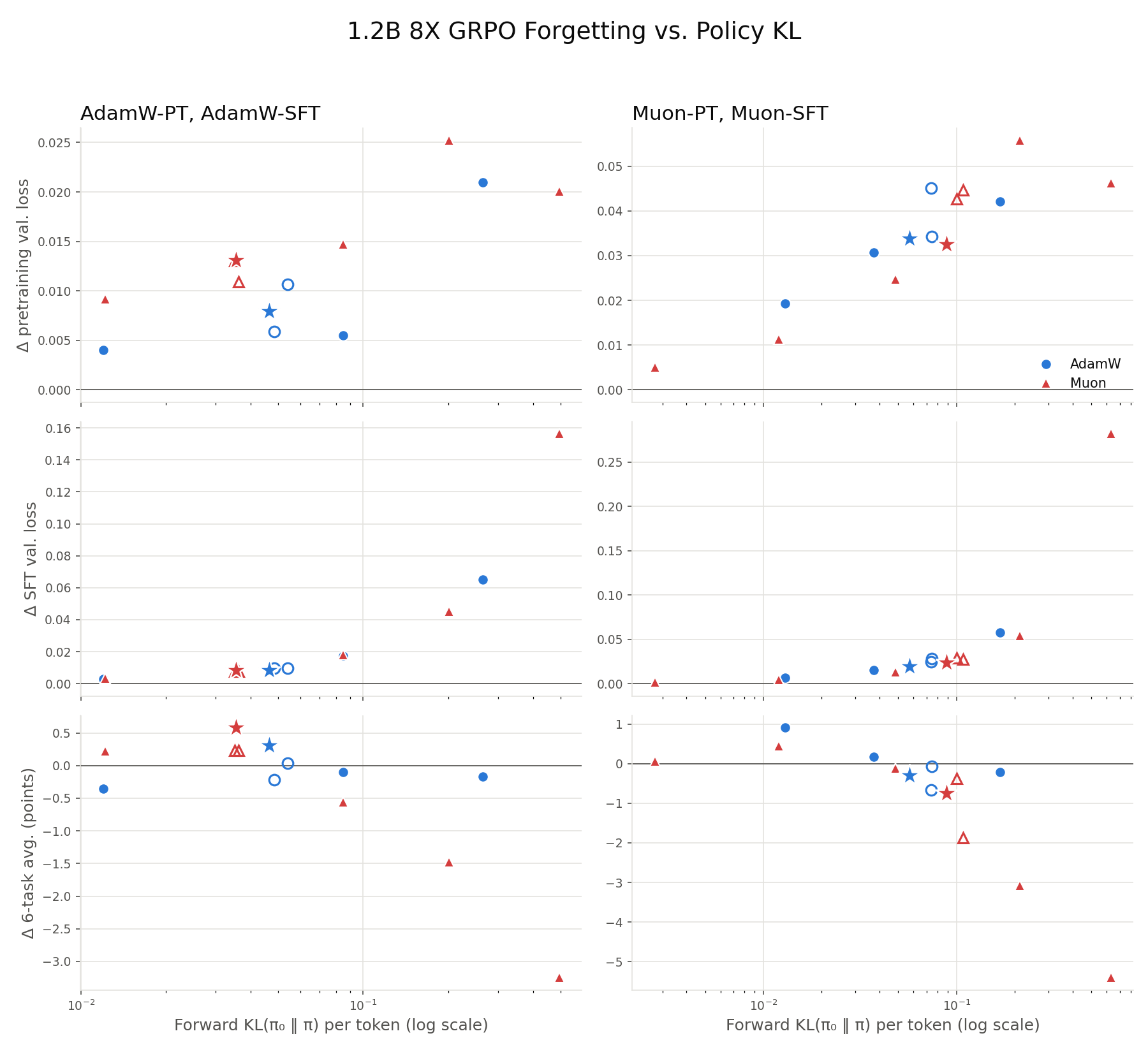}
    \caption{
        \textbf{Forgetting versus policy KL after GRPO for 1.2B models.}
        Final forgetting metrics versus forward KL from
        the initial policy across learning-rate sweeps
        for our 1.2B 8$\times$ Chinchilla models.
        Rows show changes in pretraining validation loss,
        SFT validation loss, and six-task benchmark
        average. Columns distinguish the starting models.
        Stars indicate selected learning rates based on MATH-4500 performance, hollow markers indicate other random seeds for the selected learning rates.
        Larger KL generally accompanies greater
        loss-based forgetting, but benchmark changes
        are less consistent.
    }
    \label{fig:grpo_1p2b_forgetting_vs_kl}
\end{figure}

\begin{figure}[t]
    \centering
    \includegraphics[width=1\linewidth]{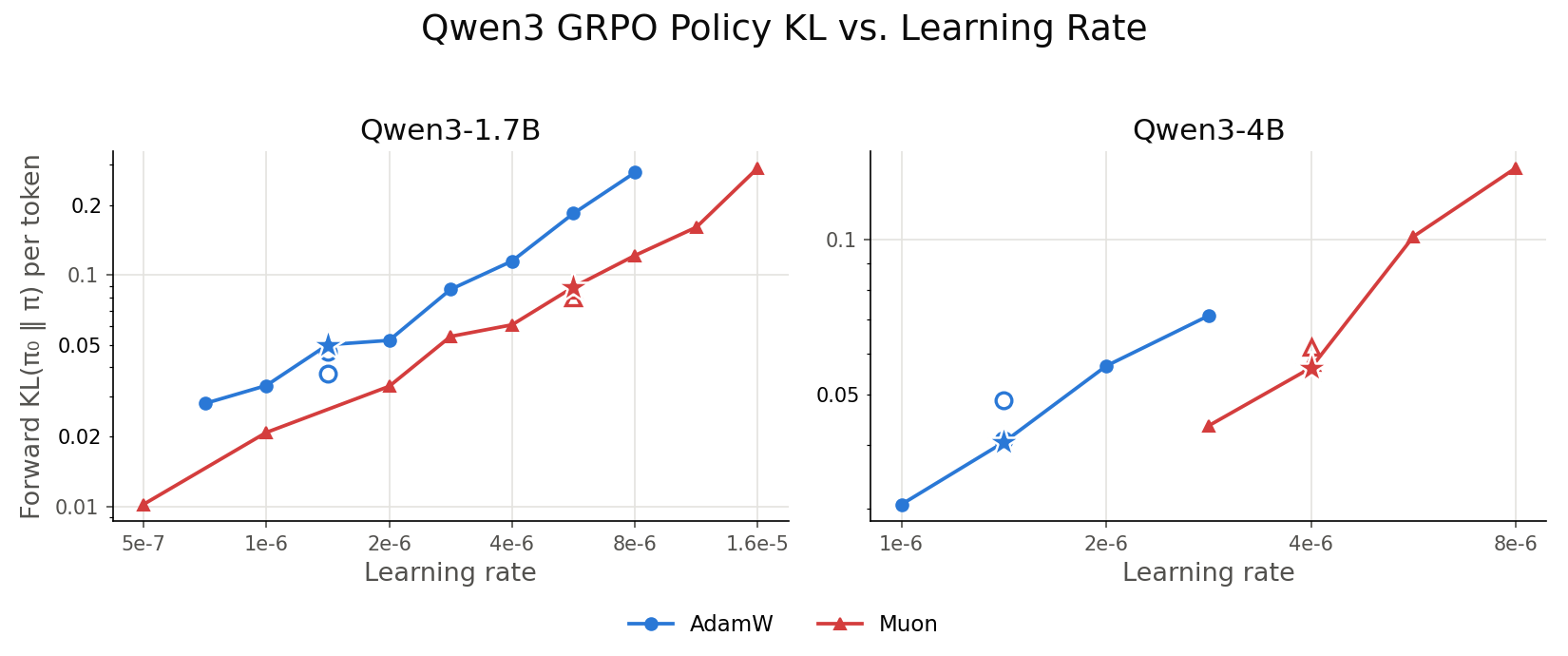}
    \caption{\textbf{Forward KL versus learning rate for Qwen3 models.}
Final forward KL from the initial policy across GRPO learning-rate sweeps for Qwen3-1.7B and 4B. Formatting follows Figure~\ref{fig:grpo_1p2b_kl_vs_lr}. As with our 1.2B models, Muon requires a higher learning rate than AdamW to achieve comparable policy drift.}
    \label{fig:grpo_qwen3_kl_vs_lr}
\end{figure}

\begin{figure}[t]
    \centering
    \includegraphics[width=1\linewidth, trim=0 0 25cm 0, clip]{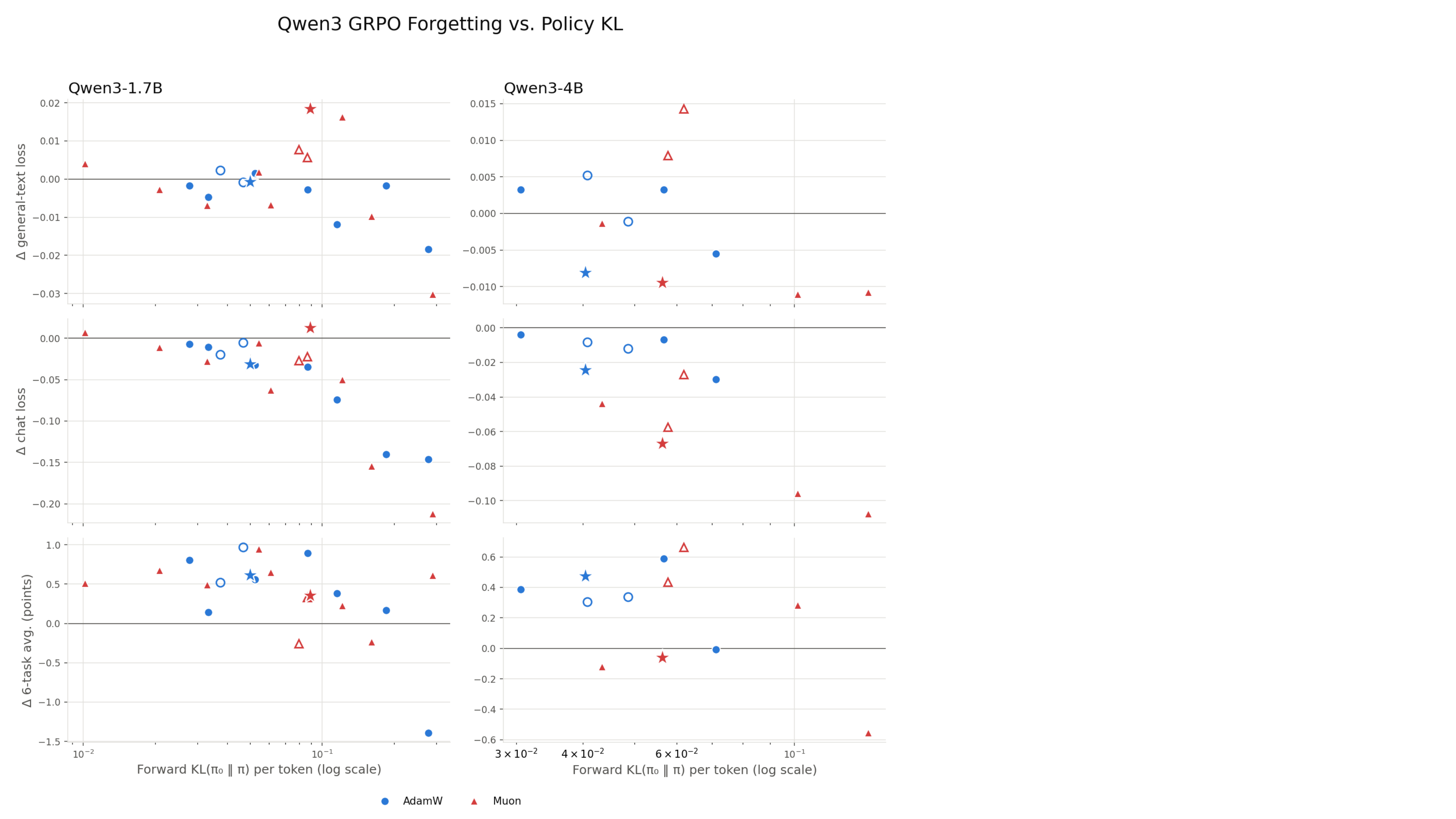}
    \caption{
        \textbf{Forgetting versus policy KL after GRPO for Qwen3 models.}
        Changes in general-text loss, chat loss, and six-task benchmark average versus final forward KL from the initial policy, across learning-rate sweeps for Qwen3-1.7B and 4B. Stars indicate selected learning rates based on MATH-4500 performance; hollow markers show additional seeds at those learning rates. We evaluate the change in loss on held-out splits of the datasets that were used for our 1.2B  pretraining and SFT stages as proxies for general-text and chat performance. Unlike the 1.2B models, the loss-based metrics show no consistent positive association with KL.
    }
    \label{fig:grpo_qwen3_forgetting_vs_kl}
\end{figure}

\paragraph{Forward KL as a predictor of forgetting.}
In Figure~\ref{fig:grpo_1p2b_traj_kl} and Figure~\ref{fig:grpo_qwen3_traj_kl}, we examine the forward KL trajectories for each optimizer at its best learning rate.
Figures~\ref{fig:grpo_1p2b_kl_vs_lr} and~\ref{fig:grpo_qwen3_kl_vs_lr} show that AdamW produces greater forward KL at the same learning rate, with Muon requiring a larger learning rate to achieve comparable policy drift. 
In Figure~\ref{fig:grpo_1p2b_forgetting_vs_kl} (1.2B models) and Figure~\ref{fig:grpo_qwen3_forgetting_vs_kl} (Qwen3 models),  we compare forgetting against forward KL across learning-rate sweeps. For our 1.2B models, both optimizers exhibit similar relationships between KL and SFT forgetting, while matching the pretraining optimizer modestly reduces pretraining forgetting at a given KL. For Qwen3, loss-based forgetting is less clearly related to KL, but larger KL generally accompanies worse six-task retention. Across all four starting models, the optimizer with lower KL at its selected learning rate also achieves better mean six-task retention. These results suggest that forward KL is a more informative measure of forgetting than learning rate or parameter-space change magnitude, despite substantial differences in the optimizers' update geometries.

\end{document}